\documentclass[smallcondensed,a4paper]{svjour3}  
\journalname{Applied Intelligence}

\usepackage{amssymb}
\usepackage{svg}
\usepackage{multicol}
\usepackage{geometry}
\usepackage{graphicx}
\usepackage{tabularx}
\usepackage{siunitx}
\usepackage{hyperref}
\usepackage{caption}
\usepackage{url}
\usepackage{amsmath}
\usepackage{booktabs}
\usepackage{ntheorem}
\usepackage{bm}
\usepackage{amssymb}

\usepackage{blindtext}
\usepackage{tikz}
\usepackage{lineno,hyperref}
\usepackage{subcaption}
\usepackage{booktabs}
\usepackage{framed}
\usepackage{etoolbox}  

\usepackage{paralist}
\AtBeginEnvironment{tabular}{\footnotesize}
\usepackage{algorithmicx,algorithm,algcompatible,float}
\AtBeginEnvironment{algorithmic}{\scriptsize} 
\algrenewcommand{\algorithmiccomment}[1]{ $\triangleright$\  #1}

\definecolor{darkgreen}{rgb}{0.0, 0.2, 0.13}

\begin{document}

\title{QCBA: Improving Rule Classifiers Learned from Quantitative Data by Recovering Information Lost by Discretisation}
\titlerunning{QCBA: Improving Rule Classifiers Learned on Prediscretised Data}
\author{Tom{\'a}\v{s} Kliegr 
\and Ebroul Izquierdo 
} 
\authorrunning{Tom{\'a}\v{s} Kliegr \and Ebroul Izquierdo } 
\institute{Department of Information and Knowledge Engineering, Faculty of Informatics and Statistics,
University of Economics Prague, W Churchill sq. 4, Prague, Czech Republic \and  School of Electronic Engineering and Computer Science, Queen Mary \\ University of London, Mile End Rd., London, United Kingdom\\
\email{first.last@\{vse.cz$|$qmul.ac.uk\}}
}

\date{Received: date / Accepted: date}

\maketitle

\begin{abstract}
A prediscretisation of numerical attributes which is required by some rule learning algorithms is a source of inefficiencies.
This paper describes new rule tuning steps that aim to recover lost information in the discretisation and new pruning techniques that may further reduce the size of rule models and improve their accuracy.  
The proposed QCBA method was initially developed to postprocess quantitative attributes in models generated by  the Classification based on associations (CBA) algorithm, but it can also be applied to the results of other rule learning approaches. We demonstrate the effectiveness on the postprocessing of models generated by five association rule classification algorithms (CBA, CMAR, CPAR, IDS, SBRL) and two first-order logic rule learners (FOIL2 and PRM). Benchmarks on 22 datasets from the UCI repository show smaller size and the overall best predictive performance for FOIL2+QCBA compared to all seven baselines.  Postoptimised CBA models have a better predictive performance  compared to the  state-of-the-art rule learner CORELS in this benchmark. The article contains an ablation study for the individual postprocessing steps and a scalability analysis on the KDD'99 Anomaly detection dataset. 
\end{abstract}
\section{Introduction}
Rules are understandable and provide a powerful means of expressing patterns present in many types of data with applications ranging from mechanical fault diagnosis \cite{fernandes2022machine} to genome analysis \cite{nawaz2021using}. With over 50 years of research (e.g., \cite{guhaOverview}), the induction of rules has been studied almost for as long as learning of neural networks (NNs). Even though it was repeatedly shown that rule models could perform on par or even better than black-box models in some tasks (e.g., \cite{rudin2019stop}), the uptake of rules remained low. For example, as of writing, no rule learning algorithm is included in the popular scikit-learn\footnote{\url{https://scikit-learn.org/}} machine learning toolkit.

One possible explanation for this is that NNs, a more successful family of machine learning algorithms, have benefited from a modular architecture that enabled many different improvements to be combined. For example, an authoritative review \cite{schmidhuber2015deep} of deep learning covers hundreds of articles presenting \emph{incremental} improvements such as  better gradient descent, the balance between complexity and high generalisation capacity, different layer architectures, etc. Throughout the same period, the progress in rule learning research did not stall, and the character of the contributions was largely different. Many rule learning papers  described a `\emph{monolithic}' method, which shared only some elementary principles with the prior art (see \cite{furnkranz2015brief} for an overview). In our perspective, the absence of a commonly recognized baseline method (such as multilayer perceptron and backpropagation in NNs), a common platform (such as TensorFlow\footnote{https://www.tensorflow.org/}) and the resulting difficulty in integrating advances is at least partly liable for the low uptake of rule learning.

In our work, we attempt to adopt the incremental paradigm, which helped to propel neural networks to their current success.  We describe multiple  postprocessing methods that yield smaller improvements when applied to models learned with multiple existing rule learners. 
As the baseline method, we use the \emph{association rule classification} group of algorithms. These provide  a key benefit: the generation of association rules is a standard task for which multiple high-performance  algorithms (and their implementations) exist.  Our implementation is for performance reasons in Java, however, as the software interface, we use the \emph{arules} ecosystem in R \cite{arules}, which enables our approach to integrate with multiple existing rule-based classifiers. A third-party Python package with scikit-learn-like interface is also available.

A major impediment of existing  association rule-based classification algorithms is that they work only on nominal data. 
Existing approaches typically address this limitation by discretisation, which is performed as part of preprocessing. For example, the ``temperature'' attribute would be automatically split into intervals such as [10,20), [20,30), $\ldots$, which are then used as independent nominal categories during rule mining. In this article, we propose several rule tuning steps aiming to recover lost information in the discretisation of quantitative attributes. We also propose a new rule pruning strategy, which further reduces the size of the classification model.

The work presented here was initially inspired by Classification based on associations (CBA)  algorithm \cite{Liu98integratingclassification}, and since the aim is to incorporate quantitative  information, we call the resulting framework ``Quantitative CBA''. The framework can also be used with other rule-based approaches that rely on prediscretised data, which we demonstrate with the case of the recently proposed Interpretable  Decision Sets (IDS) algorithm by Lakkaraju et al., 2016 \cite{lakkarajuinterpretable}, the Scalable Bayesian Rule Lists  (SBRL) algorithm by Yang et al., 2017 \cite{yang2017scalable} and four other algorithms.

The rest of this article is organised as follows: Section~\ref{sec:rw} provides a brief introduction to association rule classification.
Section~\ref{sec:marc} describes the proposed rule tuning and pruning steps. Section~\ref{sec:experiments} contains the experimental validation.
Section~\ref{ss:relatedwork} provides a comparison of the presented approach with related work and Section~\ref{sec:limitations} presents the limitations of our proposal.
The conclusions summarise the contributions. 
The Appendix
presents 
 additional  measures of quality of rule classifiers and a supplementary pseudocode listing.  

\section{Preliminaries: Association Rule Learning and Classification}
\label{sec:rw}
In this section, we introduce the association rule learning and classification tasks, and then we briefly review and compare association rule classification algorithms.
This  serves as the selection of representative algorithms that are used as baselines.

\subsection{Class Association Rule Learning}
Association rule learning  is an algorithmic approach that was originally designed to discover interesting patterns in very large and sparse instance spaces \cite{DBLP:conf/sigmod/AgrawalIS93}.  These patterns are represented as rules.

In our work, we focus on class association rule learning as a specific type of association rule learning  task, where the learned  rules are constrained to contain the target label in the consequent.
Class association rule learning is performed on similar datasets as those used for other supervised machine learning tasks. 
\begin{definition}
\textbf{(dataset)} A dataset $T$ is a set of instances. Each instance $o_i \in T$ is described by vector $\langle o_{i,1}, \ldots, o_{i,n} \rangle$, where  $o_{i,1}, \ldots, o_{i,n-1}$ are values of predictor attributes $A_1, \ldots,  A_{n-1}$, and $o_{i,n}$ is the value of the  class (target) attribute $A_n$. The value of a predictor  attribute $A$ in instance $o_i$ is denoted as $A(o_i)$ and the value of the class attribute is denoted as  $class(o_i)$.
\end{definition}

\paragraph{Example} In order to illustrate the key definitions and consequently the steps of the proposed algorithms, we will use the \texttt{humtemp} synthetic data set, denoted as $T$. There are two predictor attributes \emph{Temperature}, also referred to as $A_1$ and \emph{Humidity}, also referred to as  $A_2$. The target attribute $A_3$ is  \emph{Class} and corresponds to a subjective comfort level with values ranging from 1 (worst) to 4 (best). The dataset contains $|T|=36$ instances. An example instance $o_1 \in  T, o_1=\langle32,50,3\rangle$  corresponds to  \emph{Temperature($o_1$)}=$A_1(o_1)=32$, \emph{Humidity($o_1$)} = $A_2(o_1)=50$ and  \emph{Class($o_1$)}$=A_3(o_1)=class(o_1)=3$.     

\begin{definition}
\label{def:literal}
\textbf{(literal)} Literal  expresses the condition that an instance is set to a specific value in a given attribute. A  literal $A=V$ evaluates to true for instance $o_i$ if and only if $A(o_i) = V $, otherwise it evaluates to false. A literal $A=V$ can also be represented using the notation $A(V)$. The attribute $A$ in a literal $l=A(V)$ can also be referred to as $attr(l)$.
\end{definition}
A literal is a basic building block of association rules. It corresponds to a Boolean attribute, which is either true or false for a given instance. 
In our work, we do not consider negative literals.      

In terms of terminology, a \emph{literal} in rule learning is sometimes referred to as a \emph{condition} or as an \emph{item}. The conditions of rules in machine learning are typically represented using the $attribute=value$ notation, e.g., $A_1=32$. However, in the domain of inductive logical programming and relational rule learning, a literal is typically represented as  $attribute(value)$, e.g.,  $A_1(32)$. In the following, we adopt the former notation for practical examples of rules, and since the latter notation is more convenient for symbolic manipulations, we use it in pseudocode listings and definitions. 

\paragraph{Example} Literal $l=A_2(50)$  evaluates to true for instance $o_1$ because $A_2(o_1) = 50$ as follows from the previous example. To refer to the name of the attribute in $l$ we use $attr(l)=Humidity$.

\begin{definition}
\label{def:car}
\textbf{(class association rule)} A rule takes the form $l_1 \land l_2, \land \ldots \land l_m \rightarrow c$. The antecedent of a rule, denoted as $ant(r)$,  consists of a conjunction of literals $l_1, l_2, \ldots,  l_m$, $m \geq 0, \forall_{i,j, i \neq j}: attr(l_i) \neq attr(l_j), attr(l_i) \neq attr(c)$.    The consequent of a rule consists of one literal $c$ denoted as $cons(r)$. 
\end{definition}
A class association rule is  an implication stating that if  the conjunction of literals in the antecedent of a rule evaluates to true for an instance, then the consequent of the rule is also true for this instance.  In the rest of the article, class association rules are referred to simply as \emph{association rules} or only as \emph{rules}. For conciseness or readability reasons, the logical conjunction $\land$ is sometimes replaced by comma (,) or the word \emph{and} in rule listings.

\paragraph{Example} Consider rule
\emph{$r_1$: Temperature=32 and Humidity=50}$\rightarrow$ \emph{Class=3}. This rule contains literals $l_1=Temperature(32), l_2=Humidity(50)$ in the antecedent and one literal $l_3=Class(3)$ in the consequent. The rule meets the condition that the names of attributes in all literals are different as $attr(l_1)=$\emph{Temperature}, $attr(l_2)=$\emph{Humidity} and $attr(l_3)=$\emph{Class}.

A special type of rule is a \emph{default rule}, which is typically used as  a catch-all rule for instances that are not covered by any of the  other rules.

\begin{definition}
\label{def:rule}
\textbf{(default rule)}
If  rule $r$ is a class association rule and $ant(r)$ is empty then $r$ is a \emph{default rule}.

\end{definition}

\paragraph{Example} The rule $r_d: \emptyset \rightarrow$ \emph{Class=4} is a default rule.

When the antecedent of a rule can be applied to an instance (covers it), it does not necessarily mean that the rule correctly classifies the instance.

\begin{definition}
\textbf{(rule covering an instance)}
A rule $r$ \emph{covers}  instance $o$ if and only if all literals in the antecedent of $r$  evaluate to true for instance $o$.
If $r$ is a default rule, then $r$ covers any instance. 
\end{definition}

\paragraph{Example} Both  rules $r_1$ and $r_d$ introduced in the previous examples cover instance $o_1$. 

\begin{definition}
\textbf{(rule correctly classifying an instance)}
Rule $r$ \emph{correctly classifies} instance $o$ if and only if $r$ covers $o$ and the consequent of $r$ evaluates to true for $o$.  
\end{definition}

\paragraph{Example}
Rule $r_1$  correctly classifies instance $o_1$. Rule $r_d$ does not correctly classify instance $o$ because $cons(r_d)=4\neq3$. 

\begin{definition}\textbf{(support and confidence of a rule)}
Let $r$ be a rule, and $T$ be a dataset. 

Absolute support is computed as follows: 
\begin{equation}
supp_{abs}(r) = |\{o \in T:r \mbox{ correctly classifies } o\}|.  
\end{equation}

Support is computed as follows: 
\begin{equation}
supp(r) = \frac{supp_{abs}(r)}{|T|}.  
\end{equation}

Confidence is computed as follows:
\begin{equation}
\mbox{\emph{conf}}(r) = \frac{supp_{abs}(r)}{|\{o \in T: r \mbox{ covers  } o\}|}.  
\end{equation}

\end{definition}
Association rule learning algorithms require two parameters: minimum support and minimum confidence thresholds. Only rules with  support and confidence  equal to or above these thresholds are output. 

\paragraph{Example} Let $o_1$ be the only instance covered by $r_1$ in $T$. As shown in the previous example, instance $o_1$ is correctly classified by $r_1$, therefore $supp_{abs}(r_1)=1$. The confidence $conf(r_1) = \frac{1}{1}=1$. The support of $r_1$ in $T$ is $supp(r_1)=\frac{1}{36}=2.8\%$.

\subsection{Handling numerical data}
For numerical attributes, the standard definition of literal (Def.~\ref{def:literal}) results in problems with sparsity. For example,  single values  of temperature, such as ``Temperature=32'', could individually have too low support, hence any rule containing such a literal would not meet the prespecified support threshold.
For this reason,  quantitative attributes typically need to be discretised prior to the execution of association rule learning. The discretisation replaces the precise numerical value with an interval to increase the number of instances in data that are covered by literals. For example, when the temperature attribute is discretised, e.g., `Temperature=32' is replaced by `Temperature=(30,35]', more instances can match the latter value (interval) than the former value (specific number).

Once preprocessing has been applied, the learned rules correspond to high-density regions in  data with boundaries aligned to the discretisation breakpoints. This can impair precision but improve computational efficiency on high-dimensional data,  allowing association rule learning to  be applied to much larger data than is amenable to other types of analyses \cite[p. 492]{friedman2009elements}.

\vspace{-2mm}\subsection{Building an Association Rule Classifier} 
The first Association Rule Classification (ARC) algorithm dubbed CBA (Classification based on Associations) was introduced in 1998  \cite{Liu98integratingclassification}. While there were multiple successor algorithms, the structure of most ARC algorithms followed that of CBA  \cite{arcReview}: 
\begin{inparaenum}
\item Learn class association rules,
 \item Select the subset of the association rules,
 \item Classify new instances.
\end{inparaenum}
In the following, we briefly describe the individual steps.

\vspace{-4mm}
\subsubsection*{I. Rule Learning}
In the first step of an ARC framework, standard association rule learning algorithms are used to learn conjunctive classification rules from data.

The Apriori \cite{DBLP:conf/sigmod/AgrawalIS93} association rule learning algorithm was used in CBA, and in the recently proposed Interpretable Decision Sets (IDS). Some other approaches, such as Bayesian rule sets (BRS) \cite{wang2017bayesian}  use FP-Growth \cite{Han:2004:MFP:954514.954525}, while other algorithms such as CORELS \cite{angelino2017learning} and  Scalable Bayesian rule lists  (SBRL) \cite{yang2017scalable} are explicitly agnostic about the underlying rule learning algorithm, and could therefore be used in conjunction with any approach capable of generating regular association rules (e.g., \cite{feng2016soft,djenouri2018mining}).\footnote{The reference SBRL implementation uses ECLAT \cite{zaki2000scalable}.}

There are two main obstacles for straightforward use of the discovered association rules as a classifier: the excessive number of rules discovered even on small datasets, and the fact that the generated rules can be contradicting and incomplete (no rule covers an instance). 

\vspace{-4mm}
\subsubsection*{II. Rule selection (also called rule pruning)}
\label{ss:drpruning}
A qualitative review of rule selection (pruning) algorithms used in ARC is presented in \cite{thabtah2006pruning,arcReview}. The most commonly used method according to these surveys is \emph{data coverage pruning}.  This type of pruning processes the rules in the order of their strength, removing instances covered by the rule from the training set used for pruning. If a rule does not correctly classify at least one instance, it is deleted and the instances it covers are kept.  In CBA, data coverage pruning is combined with ``default rule pruning'': The algorithm  replaces all rules with lower precedence than the current rule with a default rule if the default rule that is inserted at that place reduces the number of errors. 
The effect of pruning on the size of the rule list is reported in \cite{Liu98integratingclassification}. Based on experiments on 26 datasets, the following effect of data coverage pruning was observed: the average number of rules per dataset without pruning was 35,140; with pruning the average number of rules was reduced to 69 without effectively impacting accuracy. 

Some ARC algorithms use optimisation algorithms to select  a subset of the candidate rules generated by association rule learning.  For example, the IDS algorithm optimises an objective function, which reflects the accuracy of the individual rules as well as multiple facets of model interpretability, including the number of rules. 

\vspace{-4mm}
\subsubsection*{III. Classification}
There are two principal types of ARC algorithms that differ in the way classification is performed: rule sets and rule lists.

\begin{definition} \textbf{(rule list)} A rule list is an ordered sequence of rules   $R=\langle r_1, \ldots, r_i, \ldots, r_m \rangle$, where the antecedents of $r_1,\ldots, r_{m-1}$ are nonempty and $r_m$ is a default rule with an empty antecedent. A rule $r_i$ has a higher precedence over  rule $r_j$ in the rule list $R$ if $i<j$ and a lower precedence if $i>j$.   An instance is classified by the highest precedence rule covering the instance.
\end{definition}

\vspace{-0.3cm}
\paragraph{Example} Example of a rule list is in Table~\ref{tbl:cbarules}.
\vspace{0.5cm}

Algorithms generating rule lists include CBA, SBRL and CORELS. While IDS uses the term ``sets'' in its title (Interpretable decision sets), the generated models are  also rule lists according to the definition presented above \cite{idsRULEML}. 
The advantage of rule lists  is that they make it easy to explain the classification of a particular instance because there is always only one rule that is accountable for it  \cite{DBLP:conf/ruleml/KliegrKSV14}.

Rule sets, also called rule ensembles, provide an alternative approach, where all rules with antecedents covering the instance are used to classify the instance. An example of an ARC algorithm combining multiple rules to perform classification is CPAR \cite{yin2003cpar}, or BRS. 

\subsection{Overview of Association Rule Classification Algorithms}
The main benefit of using a rule-based classifier, as opposed to a state-of-the-art sub-symbolic (``black-box'') method such as a deep neural network, should be the comprehensibility of the rule-based model combined with fast execution on large and sparse datasets and a competitive predictive performance.
Individual rule learning algorithms meet these aspirations to a different degree. 

\begin{table}[]
    \centering

    \begin{tabular}{p{7cm} p{6.6cm}}
        \toprule
    preferred property  & alternative \\
    \midrule
    \emph{single rule classification (rule lists)}: class prediction assigned to an instance by a rule model is attributable to a single rule & \emph{rule sets:} class is assigned by an aggregation of predictions made by an ensemble of rules \\
    \emph{crisp literals:} conditions have ``crisp'' boundaries - an instance is either covered by the rule or not & \emph{fuzzy rules:}  an instance might be covered by a rule only to a certain degree of membership\\
    \emph{deterministic learning:}  for the same training data, the result of repeated training will be the same & \emph{stochastic learning:}  a random element in the learning algorithm, such as a mutation of a rule condition \\
    \emph{small discriminative models:} rule models containing only those rules and conditions necessary to separate individual classes & \emph{larger characteristic models}: models consisting of more rules  or conditions which separate classes but also describe the predicted class\\
    \bottomrule
    \end{tabular}
    \caption{Desirable comprehensibility traits of rule models}
    \label{tab:desirable}
\end{table}
Table~\ref{tab:desirable} summarises which properties of rule models we found  desirable. In Table~\ref{tbl:comparison}, these criteria are used to compare common association rule classification algorithms. In this table, \emph{single} refers to single rule (one rule) classification, \emph{crisp} refers to whether the rules comprising the classifier are crisp (as opposed to fuzzy), \emph{det.} refers to whether the algorithm is deterministic with no random element such as genetic optimisation, \emph{assoc} corresponds to whether the method is based on association rules, \emph{acc}, \emph{rules} and \emph{time} is average accuracy, average rule count, and average time across 26 datasets as reported by \cite{alcala2011fuzzy}.

The results in Table~\ref{tbl:comparison} show that the CBA produces more comprehensible models than any of its successors considering most characteristics (single rule classification, deterministic, crisp rules). However, the CBA  tends to produce larger models than some of the newer algorithms. It should be noted that while smaller models are currently typically preferred by algorithm designers in terms of explainability, there is mixed evidence as to whether smaller models (fewer rules, fewer conditions) are always more comprehensible. Small models can be sufficient for discriminating the classes but they may not provide a sufficient explanation \cite{furnkranz2020cognitive}.

\begin{table}
{\footnotesize
\begin{center}
\begin{tabular}{p{2cm}p{0.5cm}p{0.5cm}p{0.3cm}p{0.3cm}p{0.5cm}p{0.4cm}p{0.4cm}p{1cm}}
\toprule
algorithm & year & single  & crisp  & det & assoc &  acc & \multicolumn{1}{r}{rules} &  \multicolumn{1}{r}{time} \\
\hline
CBA \cite{Liu98integratingclassification} & 1998& yes & yes & yes & yes & .80 & \multicolumn{1}{r}{185} & \multicolumn{1}{r}{35 s}\\
CBA 2 \cite{liu2001classification} & 2001 & yes & yes & yes & yes & .79 & \multicolumn{1}{r}{184} & \multicolumn{1}{r}{2 m}  \\
2SLAVE \cite{gonzalez2001selection} & 2001 & no? & no & no & no & .77 & \multicolumn{1}{r}{16} & \multicolumn{1}{r}{22 m}\\
CMAR  \cite{li2001cmar}& 2001  &  no & yes & yes & yes & .79 & \multicolumn{1}{r}{1419} &  \multicolumn{1}{r}{6 m} \\
CPAR \cite{yin2003cpar} & 2003  & no & yes & yes & yes &  .82 & \multicolumn{1}{r}{788} & \multicolumn{1}{r}{11 s}\\
LAFAR \cite{hu2003finding}& 2003 & no & no & no & yes & .75* & \multicolumn{1}{r}{47*} & \multicolumn{1}{r}{5 h*}\\
FH-GBML \cite{ishibuchi2005hybridization} & 2005  & no & no & no & no & .77 & \multicolumn{1}{r}{11} & \multicolumn{1}{r}{3 h}\\
CFAR \cite{chen2008building} &  2008 & yes & no & yes & yes & .71* & \multicolumn{1}{r}{47*} & \multicolumn{1}{r}{17 m*}\\
SGERD \cite{mansoori2008sgerd}&  2008 & no? & no & no & no & .74 & \multicolumn{1}{r}{7} & \multicolumn{1}{r}{3 s} \\
FARC-HD \cite{alcala2011fuzzy} &  2011  &no? & no& no & yes & .84 & \multicolumn{1}{r}{39} &  \multicolumn{1}{r}{1 h 20m} \\
\bottomrule
\end{tabular}
\end{center}
\vspace{-3mm}
\caption{\label{tbl:comparison} A comparison of association rule-based classifiers and closely related approaches. * indicates that the algorithm did not process all datasets}
}

\end{table}

CBA also maintains high accuracy and fast execution times.
In terms of accuracy, CBA is  outperformed only by the  FARC-HD  (by 4 percent points) and the CPAR (by 2 percent points). However, the CPAR has 4x times more rules on the output and performs multi-rule classification, which is possibly less comprehensible than the one-rule classification in CBA. While the FARC-HD outperforms CBA in terms of accuracy, this fuzzy rule learning algorithm is more than 100x slower than CBA.

It should be noted that Table~\ref{tbl:comparison}, which is based on data from \cite{alcala2011fuzzy}, excludes several relevant recently proposed algorithms. These typically subject the input rule set generated by association rule learning to a sophisticated selection process, involving optimisation techniques such as the Markov chain Monte Carlo (in SBRL), branch-and-bounds (in CORELS), submodular optimisation (in IDS) or  simulated annealing (in BRS). For these algorithms, a comprehensive previously published benchmark on a larger number of datasets is not,  to our knowledge, available. Nevertheless, these algorithms are widely considered state-of-the-art in the area of association rule classification (e.g. in \cite{wang2018multi}). We select SBRL and IDS as two additional algorithms, which are postprocessed with our approach and evaluated. CORELS is also included among the benchmarked algorithms.
\vspace{-4mm}

 \floatname{algorithm}{Procedure}
\begin{algorithm}[h] 
\caption{QCBA \emph{qcba().}}
\label{alg:qcba}
{\tiny
\begin{algorithmic}[1]
\REQUIRE $rules$ input rule set (order not important)
\ENSURE $rules$ ordered rule list
\STATEx
\STATE  $ rules \leftarrow$ remove rules with empty antecedent from $rules$.
\FOR{$rule \in rules$ } \Comment{Can be parallelized}
\STATE  $ rule \leftarrow refit(rule)$ \Comment{see Procedure~\ref{alg:refit}} 
\STATE  $ rule \leftarrow pruneLiterals(rule)$\Comment{see  Procedure~\ref{alg:attPruning}} 
\STATE  $ rule \leftarrow trim(rule) $\Comment{see Procedure~\ref{alg:trim}}
\STATE $ rule \leftarrow  extendRule(rule)$ \Comment{see Procedure~\ref{alg:enlarge} - \ref{alg:extension}}
\ENDFOR
\STATE \emph{rules} $\leftarrow postprune(rules)$ \Comment{see Procedure.~\ref{alg:databasecoverage}, post-pruning adds a new default rule, if post-pruning is not performed, the QCBA ensures that default rule is added/updated and sorted.}
\STATE \emph{rules} $\leftarrow drop(rules)$ \Comment{Proc.~\ref{alg:defRuleOverlapTrans} for instance-based  and  Proc.~\ref{alg:defRuleOverlapRange} for range-based version of default rule overlap pruning} 
\STATE \textbf{return} {$rules$}
\end{algorithmic}
}
\end{algorithm}
\vspace{-10mm}
\section{Proposed Approach}
\label{sec:marc}
Quantitative classification based on associations (QCBA) is a collection of rule tuning  steps that use the original continuous data to ``edit'' the rules, changing the scope of literals  in the antecedent of the rules. As a consequence, the fit of the individual rules to the data improves. The second group of tuning steps aims to remove unnecessary  literals and rules, making the classifier smaller and possibly more accurate. The resulting models are ordered rule lists with favourable comprehensibility properties, such as one-rule classification and crisp rules. 

The method takes on the input as follows:
\begin{itemize}
  \item One input is training dataset $T$ with numeric attributes (before discretisation) used to learn the rules.
 \item Another input is the set of rules learned on a discretised version of $T$.
\end{itemize}
The input to the algorithm is an unordered set of rules, but it can also be an ordered rule list. In the latter case, the input order is ignored.  
The output is an ordered classification rule list, which has the same or a smaller number of rules (a default rule may be added). The individual rules have the boundaries of numerical conditions (literals) adjusted. Some may also be completely removed. The first phase consists of the following rule tuning (optimisation\footnote{We prefer to use the term ``tuning'', because the term ``optimisation'', which is used in the same context in some other papers, such as \cite{huhn2009furia}, evokes the application of mathematical programming. }) 
steps:

\begin{enumerate}
\item \textbf{Refitting rules to the value grid.} Literals initially aligned to the borders of the discretised  regions are refit to a finer grid with steps corresponding to all unique attribute values appearing in the training data.
\item \textbf{Literal pruning}: Redundant literals are removed from the rules.
\item \textbf{Trimming.} Boundary regions with no correctly classified instances are removed.
\item \textbf{Extension.} Ranges of literals in the antecedents are extended, preserving or improving rule confidence. 
\end{enumerate}

Once the rules have been processed, their coverage changes which can make some of the rules redundant.  

\begin{enumerate}
 \setcounter{enumi}{4} 
\item \textbf{Data coverage pruning and default rule pruning.} These two pruning steps, proposed in \cite{Liu98integratingclassification}, correspond to the classifier builder phase of CBA.  
\item \textbf{Default rule overlap pruning.} Rules that classify into the same class as the default rule at the end of the classifier can be removed if there is no other rule between the removed rule and the default rule that would change the classification of instances initially classified by the removed rule. 
\end{enumerate}

Procedure~\ref{alg:qcba} depicts the succession of tuning steps in the QCBA and provides pointers to the algorithms described in detail in the following subsections. For each proposed step, there is a brief description of the effects on selected measures of classifier quality measures. For tuning steps 1--4, these are measures applying to single rules as ``local classifiers''. Pruning steps 5--6 are evaluated in terms of their effect on the complete rule list -- the ``global classifier''. 

 \begin{figure}
         \centering
         \includegraphics[width=\textwidth]{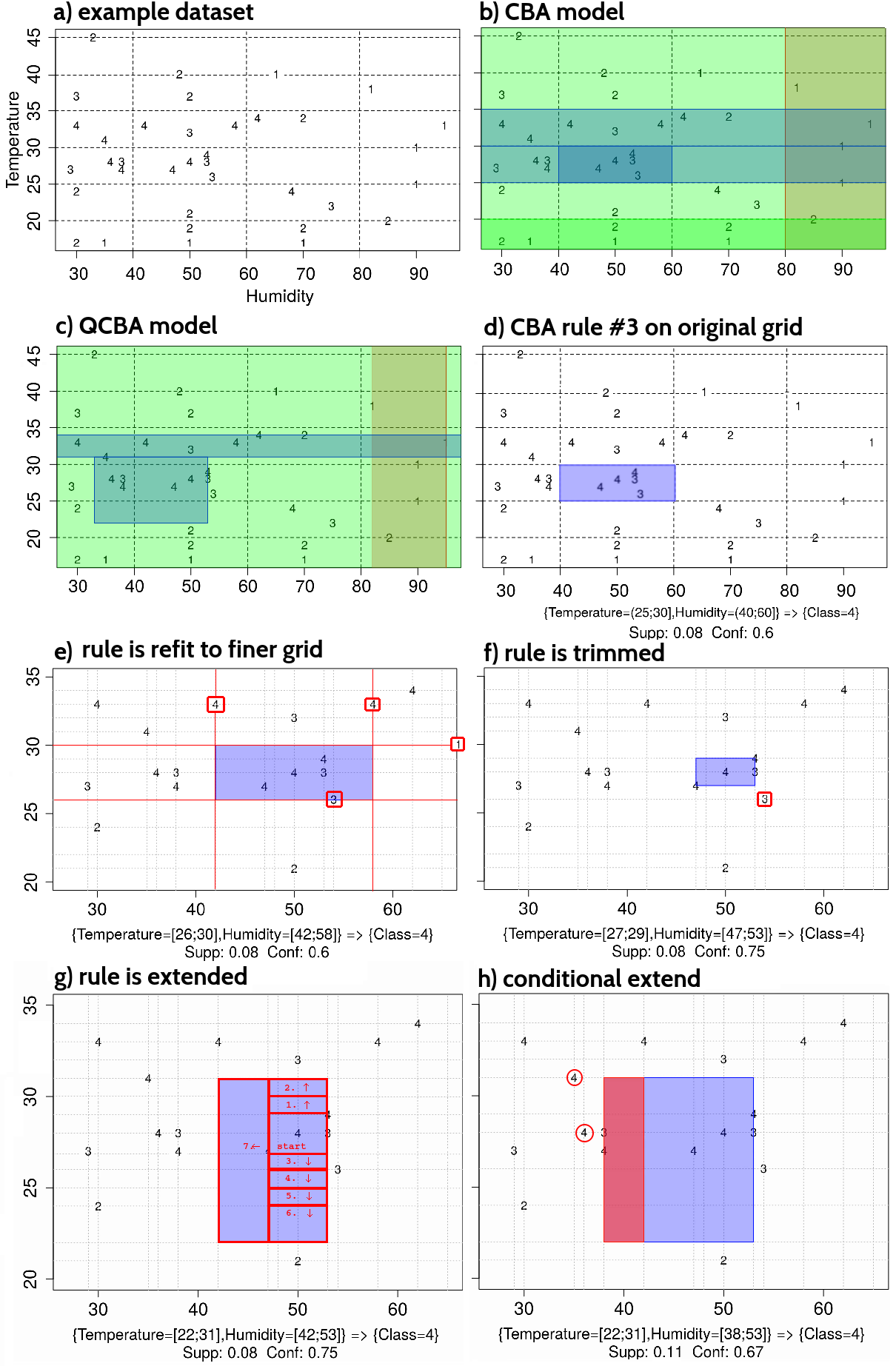}
     \caption{Illustration of refit, trimming and extension tuning steps. 
     }
     \label{fig:all}
 \end{figure}
\vspace{-2mm}\subsection{Example}
\label{sec:example}
To illustrate the rule tuning steps, we use the \texttt{humtemp} synthetic dataset. The dataset is plotted in Figure~\ref{fig:all}a. 

The grid depicted with the dashed lines denotes the result of a discretisation algorithm, which is performed as part of preprocessing. In this case, equidistant binning  is applied. Figure~\ref{fig:all}b and Table~\ref{tbl:cbarules} show an input list of rules learned on the \texttt{humtemp} data. In our example, these rules are learned by CBA, but another algorithm generating rule lists could have also been applied.

A QCBA model generated after all of the tuning steps is shown in Fig.~\ref{fig:all}c.
Fig.~\ref{fig:all}e-h correspond to the individual tuning steps, which transform the original rule R\#3 (see Figure~\ref{fig:all}d, Table~\ref{tbl:cbarules}) from the CBA model  to its final form in the QCBA model. These figures will be referred to again from the detailed description included below.

\begin{table}[h]
\centering
 \begin{tabular}{lllll}
 \toprule
 id & antecedent & consequent & supp & conf \\
 \midrule
 1 & Humidity=(80,100]     &              $\rightarrow$ {\color{red} Class=1}  & 0.11  & 0.80 \\
 2  &Temperature=(30,35]      &            $\rightarrow$ {\color{blue} Class=4}  & 0.14 & 0.63 \\
 3 &Temperature=(25,30] and Humidity=(40,60]  &  $\rightarrow$ {\color{blue}  Class=4} & 0.08 & 0.60 \\
 4 &Temperature=(15,20]    &              $\rightarrow$ {\color{darkgreen}  Class=2} & 0.11 & 0.57\\
 5 &Temperature=(25,30]      &         $\rightarrow$ {\color{blue}  Class=4} & 0.14 & 0.50 \\
 6 &  & $\rightarrow$ {\color{darkgreen} Class=2}                                     & 0.28 &  0.28\\
 \bottomrule
 \end{tabular}
 \caption{Input rule list depicted in Figure~\ref{fig:all}b. Predicted classes are colour-coded.}
  \label{tbl:cbarules}
\end{table}

\vspace{-2mm}\subsection{Refit}
The refit step is inspired by the way the C4.5 decision tree learning algorithm \cite{DBLP:books/mk/Quinlan93} selects splitting points for numerical attributes. 
Fig.~\ref{fig:all}d shows rule ``Rule \#3: Temperature=(25,30] and Humidity=(40,60]\hspace{0.1cm}  $=>$ Class=4'' contained in the CBA model. Note the ``padding'' between the rule boundary and the covered instances that are nearest to the boundary.

 \begin{algorithm}[H] 

  \scriptsize
 \caption{Refit rule \emph{refit()}}
 \label{alg:refit}
 \begin{algorithmic}[1]
 
 \REQUIRE $r$ input rule, $T$ dataset
 \ENSURE rule  $r$ with refit literals
 \STATEx
 \FOR{literal $l=A(I) \in  ant(r)$} \COMMENT{Non-interval valued literals are skipped}
 
 \STATE $I' \leftarrow [min(vals(l,T)), max(vals(l,T))]$ \Comment{ where vals(l,T) is a value set, see  Def.~\ref{def:vr}}
 \STATE $r$ $\leftarrow$ replace literal $A(I)$ in $r$ with $A(I')$
 \ENDFOR
\STATE \textbf{return} {$r$}
 \end{algorithmic}
 \end{algorithm}

The refit tuning step (Procedure~\ref{alg:refit})  contracts interval boundaries to a finer grid, which corresponds to the raw, ``undiscretised'' attribute values ensuring that the refit rule covers the same instances as the original rule. 
The standard definition of a literal truth value (Def.~\ref{def:literal}) requires that the value  in the literal exactly matches the value of the referenced attribute in the instance.  To do this, we need to amend this definition to also apply to the values in the training data that belong to the interval. 

\begin{definition}
\label{def:intervalvaluedlit}
\textbf{(interval-valued literal)} Literal $A(I)$ is called an interval-valued literal if  $I$ is an interval. An interval-valued literal $l$ evaluates to true for instance $o$ if and only if $ A(o)$ belongs to the interval $I$. 
\end{definition}

\paragraph{Example} 
Literal \emph{Temperature}$=(25,30]$ is true according to Def.~\ref{def:literal} only for instances that have the exact string value `$(25,30]$' in the Temperature attribute.
Since the QCBA works with the original values before discretisation,  according to Definition~\ref{def:intervalvaluedlit}, the literal \emph{Temperature}$=(25,30]$ is also true for an instance with the numeric value 26 in the \emph{Temperature} attribute.

\begin{definition}
\label{def:vr}
\textbf{(value set of a literal)}
We let $l=A(V)$ be a literal defined on attribute $A$ and  $T$ be a dataset. Then we let the value set of $l$  be defined as  $vals(l,T)=\{A(o):o \in T, l \mbox{ evaluates to true for } o\}$.

\end{definition}

\paragraph{Example}  For  dataset $T$ depicted in Figure~\ref{fig:all}a,  the value set of literal  $l:$ \emph{Temperature}$=(25,30]$ is  the set of five temperature readings that are actually recorded in the dataset: $vals(l,T)=\{26,27,28,29,30\}$.

The Refit step of the proposed QCBA method needs to retrieve the source values that are merged into an interval. These are not defined as all real numbers included in the interval, but rather as values actually appearing in the referenced attribute in the instances in a supplied dataset.  
This definition will be referenced from the extension tuning step.

\begin{definition}
\label{def:ar}
\textbf{(value set of an attribute)}
We let the value set of attribute $A$ in dataset $T$   be defined as  $vals(A,T)=\{A(o):o \in T\}$.
\end{definition}

\paragraph{Example}
The red boxes in Figure~\ref{fig:all}e mark the instances that are used as ``anchors'' for the refit of Rule \#3 from Table~\ref{tbl:cbarules}. For the literal ``Temperature=(25,30]'', the upper boundary corresponds to an existing instance; therefore there is no change. Since the lower boundary is exclusive, it is adjusted to the nearest value of a real instance, which is 26.  Likewise, the boundaries of Humidity in Rule \#3 are adjusted to values of the nearest instances within the original boundaries. The resulting rule is ``Temperature=[26,30],  Humidity=[42,58] $\rightarrow$ Class=4'' shown in Figure~\ref{fig:all}e.

\emph{Effect on measures of classifier quality}  Since refit does not affect the covered or classified instances on the training dataset, the confidence and the support of the rule after refit are unchanged (as computed from the training data). The main effect of the refit step is the increase in the density of the region covered by the rule: the same instances as covered by the original rule are covered by a smaller region corresponding to the refitted rule. \emph{Density} is  computed as the number of correctly classified instances divided by the volume covered by the antecedent of the rule (see Definitions~\ref{def:rulevolume},\ref{def:ruledense} in Appendix~\ref{sec:properties}).

\vspace{-2mm}\subsection{Literal pruning}
Some association rule learning algorithms can output models containing rules with redundant literals. One example is IDS, where rules are generated without checking the confidence threshold. Since the IDS algorithm contains a stochastic element, there is no guarantee that even if a shorter rule covering the same instances is included among the candidate rules, it will be selected into the final classifier instead of the longer rule.  As a consequence, the classifier can contain rules with unnecessary literals.

The applicability of literal pruning is not limited to IDS. Since QCBA is modular, literal pruning can also be used after other steps in QCBA have been performed. For example, while candidate rules used in CBA do not generally contain redundant conditions, after the literals in these rules have been subject to other QCBA tuning steps, especially trimming and extension, some of the conditions may become redundant.

The proposed literal pruning algorithm attempts to remove literals from a rule. If the confidence of the rule does not drop, the shorter rule is kept and becomes a seed for further attempts at literal pruning.
Literal pruning is depicted in Procedure~\ref{alg:attPruning}.

 \begin{algorithm}[H] 
 \caption{Literal pruning \emph{pruneLiterals()}}
 \label{alg:attPruning}
 \begin{algorithmic}[1]
 \REQUIRE $r$  input rule 
 
 \ENSURE rule $r$ with redundant attributes (literals) removed
 \STATEx
 \REPEAT
  \STATE \emph{removed} $\leftarrow$ \emph{false}
 \FOR{$literal \in  ant(r)$} \Comment{Literals are iterated in the order of appearance in the input rule}
   \STATE  $r' \leftarrow$ remove $literal$ from $r$ 
   \IF{\emph{conf(}$r'$\emph{)} $\geq$ \emph{conf(r)}}
     \STATE $r \leftarrow r'$; \emph{removed} $\leftarrow$ \emph{true}
     \STATE \textbf{break}
   \ENDIF
 \ENDFOR
 \UNTIL{\textbf{not} \emph{removed}}
 \STATE \textbf{return} {$r$}
 \end{algorithmic}
\end{algorithm}
In our experiments, it can be seen that from most rules no literal can be pruned. Therefore iteration in the order of appearance is more computationally efficient  than a full greedy procedure that would evaluate all candidate literals for removal and then select the best one.  Additionally, literal pruning can result in duplicate rules in the rule list. These are removed as part of the post-pruning step.

\emph{Effect on measures of classifier quality}
Literal pruning improves (or does not affect) the quality measures of individual rules as local classifiers, as rule confidence, and rule support can only stay the same or increase (on the training data). The rule length stays the same or decreases, making the rules shorter.    As each omitted condition means that the rule spans  all values indiscriminately in the corresponding attribute, the density of instances covered by the rule decreases when one or more literals are removed.

\vspace{-5mm}\subsection{Trimming}
This step adjusts the interval boundaries to the actual values in the covered instances.
If there is any covered but misclassified instance on the boundary, it is removed, the boundary is  adjusted, and this process repeats until no instance is removed (Procedure~\ref{alg:trim}).

 \begin{algorithm}[H] 
 \caption{Rule trimming \emph{trim()}}
 \label{alg:trim}
 \begin{algorithmic}[1]
 \REQUIRE $r$ input rule, $T$ dataset 
 \ENSURE  rule $r$ with trimmed literals
 \STATEx
 \STATE $T_{r}^{corr} \leftarrow \{o\in T, o  \mbox{ correctly classified by } r\}$
 \FOR{$l=A(I) \in  ant(r)  }$ \Comment{Literals are iterated in the order of appearance in the rule, non-interval valued literals are skipped}
 \IF{$|vals(l,T_{r}^{corr})| \leq 1$} \COMMENT{Interval containing a single  value or empty interval.}
 \STATE \textbf{continue} 
\COMMENT{These are skipped because trimming could create a condition with empty value set}
 \ENDIF
     \STATE $I' \leftarrow  [min(vals(l,T_{r}^{corr})),max(vals(l,T_{r}^{corr}))] $   
     \STATE $r \leftarrow$ replace literal $l$ in $r$ with literal $l'=(A,I')$
 \ENDFOR
 \STATE \textbf{return} {$r$}
 \end{algorithmic}
 \end{algorithm}

\paragraph{Example} In Figure~\ref{fig:all}f, trimming has been applied and the rule is shaved of boundary regions that do not cover any correctly classified instances: one instance with class 3 (denoted by a red box) on the rule boundary is initially covered by Rule \#3, but also misclassified by it. As part of the trimming, the lower boundary of the $Temperature$ literal on Rule \#3 is increased not to cover this instance.

\emph{Effect on measures of classifier quality}
Since only regions with incorrectly classified instances are removed, the trim operation does not change the  support of the trimmed rule, and rule confidence can only increase or stay the same. The density of the region covered by the rule is also increased (or unchanged), since the region covered by the rule stays the same or decreases, and the number of correctly classified instances stays the same.

\vspace{-2mm}\subsection{Extension}
\label{ss:Ext}
The purpose of this hill-climbing process is to move literal boundaries outwards, increasing rule coverage. 
Each extension generated by Procedure~\ref{alg:enlarge} corresponds to a new rule that is derived from the current rule by extending the boundaries of literals in its antecedent in one direction with steps corresponding to breakpoints on the finer grid.

\begin{algorithm}
\caption{Rule Extension \emph{getRuleExtension()}}
\label{alg:enlarge}
\begin{algorithmic}[1]
\REQUIRE $r$ rule to be extended, \emph{l} literal on which extension should be performed, \emph{type} of extension 
\ENSURE  extension of rule $r$ is a tuple (extended rule, extended literal, direction of the last performed extension) or an empty set
\STATEx

  \STATE $l_e \leftarrow$ direct extension of $l$  of specified $type$ \COMMENT{Def.~\ref{def:dirExtCard}} 
  \IF{$l_e \neq \emptyset$}
  \STATE $r  \leftarrow$ replace $l$ in $r$ with $l_e$
\STATE \textbf{return }  \emph{$(r,l_e,type)$}
\ELSE
\STATE \textbf{return} $ \emptyset$
\ENDIF
\end{algorithmic}
\end{algorithm}

The algorithm refers to the notion of the direct extension of a literal presented Definition~\ref{def:dirExtCard}.
\begin{definition} 
\label{def:dirExtCard}
\textbf{(direct extension of an interval-valued literal)} We let $l=A(][x_i,x_j][)$ be an interval-valued literal and a $T$ a dataset, where ``]['' denotes that the interval boundary is either open our closed.
 A \emph{direct extension} of type \emph{lower} of literal $l$ is a literal $A([x_l,x_j][)$, where the lower boundary is $x_l=  max(\{x\in vals(A,T): x < min(vals(l,T)) \})$ and the type of the upper boundary (open or closed) is the same as in $l$. A \emph{direct extension} of type \emph{higher} of literal $l$ is a literal $A(][x_i,x_h])$, where the upper boundary is  $x_h=  min(\{x\in  vals(A,T): x > max(vals(l,T)) \})$  and the type of the lower boundary is the same as in $l$. If $x_l=\emptyset$ then a direct extension of type \emph{lower} does not exist (is empty set). If $x_h=\emptyset$ then the direct extension of type \emph{higher} does not exist (is empty set). 

\end{definition}

 The extension process for a given rule can generate multiple extension candidates (Procedure~\ref{alg:neighbourhood}).

\begin{algorithm}
\caption{Get All Direct Candidate Extensions \emph{getAllDirectExtensions()}}
\label{alg:neighbourhood}
\begin{algorithmic}[1]
  \REQUIRE
$r$ rule for which direct extension candidates should be generated 
\ENSURE  set of extension candidates for $r$ consisting of tuples (extended rule, extended literal, direction of the last performed extension) or an empty set
\STATEx
\STATE $R_e$ $\leftarrow \emptyset$
\FOR{$l  \in  ant(r)$} \COMMENT{Process only interval-valued literals in the order of appearance in the rule} \label{l:extit}
\FOR{$type \in \{lower,higher\}$}
  \STATE $R_e \leftarrow R_e \cup getRuleExtension(r,l,type)$\COMMENT{see  Proc. \ref{alg:enlarge}}
  \ENDFOR
\ENDFOR
\STATE \textbf{return } $R_e$
\end{algorithmic}
\end{algorithm}

\begin{algorithm}
\caption{Rule Extension \emph{extendRule()}}
\label{alg:extension}
\begin{algorithmic}[1]
\REQUIRE $r$ rule to be extended, \emph{minImprovement} $\in (-1,1)$ with 0 as default,  $minCI$ $\in [-1,0)$ with -1 as default
\ENSURE extended rule $r$
\STATEx
\STATE  $r_{best} \leftarrow r$
  \REPEAT
    \STATE $E$ $\leftarrow$ $getAllDirectExtensions$($r_{best}$) \label{algLine:extensionCand}\COMMENT{see  Proc.~\ref{alg:neighbourhood}} 
    \STATE \emph{extensionSuccessful} $\leftarrow $ \textbf{false}
    \FOR{$(r_e,l_e,type_e)  \in  E  $ \COMMENT{Iteration in order according to criteria in Def.~\ref{def:ranking}}}
      \STATE $\Delta_{\mbox{\emph{conf}}} \leftarrow$  conf($r_e$) - conf($r_{best}$)
      \IF{$\Delta_{\mbox{\emph{conf}}} \geq minImprovement$}        \COMMENT{Crisp acceptance}     \label{algLine:extension:firstcheck}
        \STATE $r_{best}  \leftarrow r_e$
        \STATE \emph{extensionSuccessful} $\leftarrow $ \textbf{true}
        \STATE \textbf{break}
       \ENDIF
 \STATE 
        \WHILE{\textbf{not}(\emph{extensionSuccessful}) \textbf{and} $\Delta_{\mbox{\emph{conf}}} \geq minCI$  }         \Comment{Conditional acceptance} \label{line:condaccept}
            \label{algLine:extension:condExtensionLoopStart}    
            \STATE $(r_e,l_e,type_e) \leftarrow getRuleExtension(r_e, l_e, type_e)$ \COMMENT{Proc. ~\ref{alg:enlarge}}
            \IF{$(r_e,l_e,type_e) = \emptyset$}
            \STATE \textbf{break} 
            \ENDIF 
            \STATE $\Delta_{\mbox{\emph{conf}}} \leftarrow conf(r_e) - conf(r_{best})$ 
            \IF{$\Delta_{\mbox{\emph{conf}}} \geq minImprovement$} \label{algLine:extension:secondcheck}
                \STATE $r_{best}  \leftarrow r_e$,  \emph{extensionSuccessful} $\leftarrow $ \textbf{true}, $E\leftarrow \emptyset$
            \ENDIF
	    \ENDWHILE
\label{algLine:extension:condExtensionLoopEnd}
    \ENDFOR
  \UNTIL{\emph{extensionSuccessful} = \textbf{false}}
\STATE  \textbf{return} $r_{best}$
\end{algorithmic}
\end{algorithm}

In Procedure~\ref{alg:extension}, the candidate extensions are iterated in the order specified by CBA: 
\begin{definition} \textbf{(rule sorting criteria)}
  The rule $r_i$ is ranked higher than the rule $r_j$ if
\begin{enumerate}
\item  \emph{conf}$(r_i) > $ \emph{conf}$(r_j)$ 	 \hspace{0.5cm } $\triangleright$ higher confidence rules are preferred,
\item $supp(r_i) > supp(r_j)$   \hspace{0.5cm } $\triangleright$ higher support rules are preferred,
\item $|ant(r_i)| < |ant(r_j)|$  \hspace{0.5cm } $\triangleright$ shorter rules are preferred.
\end{enumerate}
\label{def:ranking}
The first applicable condition is used.
\end{definition}

An extension is accepted if there is an improvement in confidence over the last confirmed extension by a user-set $minImprovement$ threshold. If this  \emph{crisp extension} has not been reached, a \emph{conditional extension} is attempted  (Procedure~\ref{alg:extension}, line~\ref{line:condaccept}). This picks up on the last extension attempt that resulted in a rule that does not meet the criteria for crisp extension but it meets the criteria for the conditional extension. The procedure iteratively generates new candidate rules by extending  a literal that was subject to extension in this last extension attempt. If the extension meets the criteria for crisp acceptance, the conditional extension successfully finishes with the extension replacing the current rule. 

\paragraph{Example} The red boxes in Fig.~\ref{fig:all}g demonstrate the confirmed extension steps for the seed rule depicted in Fig.~\ref{fig:all}f. In the first confirmed extension, the upper \emph{Temperature} boundary is increased to 30 by one step of the finer grid. Since the confidence of the extended rule does not decrease, this extension meets the conditions for crisp acceptance.
Subsequent extensions result in the rule depicted by the blue region in Fig.~\ref{fig:all}g. This rule cannot be further extended in any of the directions without a drop in confidence. The procedure, therefore, tries  the conditional extension (red region in Fig.~\ref{fig:all}h) by lowering the boundary of the \emph{Humidity} literal to 38. This makes the rule cover two more instances, one correctly and one incorrectly, decreasing the confidence below the initial 75\%. 
Further extends in \emph{Humidity} in the same direction (by lowering the lower boundary) make the rule cover two more instances correctly (these are in red circles in Fig.~\ref{fig:all}h). This brings the confidence back to 75\% and  results in  a crisp acceptance. While the algorithm tries the additional conditional extensions, none were successful. Fig.~\ref{fig:all}c shows the final rule, which has both a higher confidence and support than the original rule in Fig.\ref{fig:all}d.

\emph{Effect on measures of classifier quality}
Assuming $minImprovement=0$ (the default setting), the procedure outputs only rules with confidence higher or the same as the original rule. Since the algorithm does not have any ``shrink'' step, as it can only extend the coverage of the rule by adding new regions and instances, the support of the rule can only increase or remain unchanged.
The density of the rule can increase, decrease or stay the same as a result of extension depending on the ratio of the increase in the number of correctly classified instances and the size of the newly covered regions as a result of the extension.

\vspace{-2mm}\subsection{Post-pruning}
The previous steps affected individual rules, changing their coverage. The number of rules can now be reduced using an adaptation of CBA's data coverage pruning and the default rule pruning. The latter also adds a default rule to the end of the rule list that ensures that the rule list covers all possible test instances.
As a first step of post-pruning, rules are sorted according to the criteria in Definition~\ref{def:ranking}.
The post-pruning algorithm is depicted in Procedure~\ref{alg:databasecoverage}  and corresponds to the core of the CBA algorithm described in Section~\ref{sec:rw}.

 \begin{algorithm}
 \caption{post-pruning \emph{post-pruning()}}
 \label{alg:databasecoverage}
 \begin{algorithmic}[1]
 \REQUIRE \emph{rules} to be pruned with exactly one default rule, $T$ training dataset with class attribute $C$
 \ENSURE pruned rules (some elements of $rules$ removed) and default rule added or updated 
 \STATEx 
 \STATE $rules \leftarrow \{r \in rules: |ant(r)|>0\}$ \COMMENT{remove any default rule}
 \STATE \emph{cutoffRule} $\leftarrow \emptyset$
 \STATE \emph{defClass}$\leftarrow$  $arg max_{cl\in   vals(C,T)} |\{o\in T: class(o)=cl\}|$ \Comment{most frequent class in training data}

 \STATE \emph{defaultRuleError} $\leftarrow |T| -|\{o\in T: class(o)=\mbox{\emph{defClass}}\}|$
 \STATE \emph{lowestTotalError} $\leftarrow $\emph{defaultRuleError}  
  \STATE \emph{cutoffClass} $\leftarrow$ \emph{defClass}
 \STATE \emph{totalErrorWithoutDefault} $\leftarrow 0$
     \STATE  \Comment{Data coverage pruning}
     \FORALL{\emph{r} $\in$ \emph{rules}}\Comment{Process rules in the order given by Def.~\ref{def:ranking}}
     \STATE  \emph{covered} $\leftarrow \{o \in \emph{T}: o \mbox{ covered by } r\}$
      \STATE  \emph{corr} $\leftarrow \{o \in \emph{T}: o \mbox{ correctly classified by } r\}$
       \IF{\emph{corr}=$\emptyset$}
       \STATE $rules$ $\leftarrow rules \setminus r$ \Comment{remove \emph{r} from \emph{rules}}
       \ELSE      
             \STATE  $T  \leftarrow T \setminus covered$ \Comment{Remove instances covered by $r$ from training data}
 	\STATE \emph{misclassified} $\leftarrow |\emph{covered}| - |\emph{corr}|$
 	\STATE \emph{defClass} $\leftarrow arg max_{cl\in   vals(C,T)} |\{o\in T: class(o)=cl\}|$ 
 	\STATE \emph{totalErrorWithoutDefault} $\leftarrow$ \emph{totalErrorWithoutDefault} +\emph{ misclassified}
         \STATE \emph{defaultRuleError} $\leftarrow |T| - |\{o\in T: class(o)=\mbox{\emph{defClass}}\}|$    
         \STATE \emph{totalErrorWithDefault} $\leftarrow$ \emph{defaultRuleError} + \emph{totalErrorWithoutDefault}
       \IF{\emph{totalErrorWithDefault}$<$\emph{lowestTotalError}}
       \STATE \emph{cutoffRule, lowestTotalError, cutoffClass} $\leftarrow r,  totalErrorWithDefault, defClass$
       \ENDIF
       \ENDIF
             
     \ENDFOR
     \STATE \Comment{Default rule pruning}
     \STATE $rules$ $\leftarrow$ remove all rules with lower precedence than \emph{cutoffRule}  from $rules$
     \STATE  $rules$ $\leftarrow$ append new default rule  ``$\emptyset \rightarrow $ \emph{C=cutoffClass}'' at the  end of $rules$ (at position with the lowest precedence)
     \STATE  \textbf{return}  \emph{rules}
 
 \end{algorithmic}
 \end{algorithm}

\emph{Effect on measures of classifier quality}
Unlike all the preceding steps, the standard CBA data coverage pruning does not edit the individual rules, but by selecting a subset of rules, it forms the final ``global'' classifier.
The data coverage pruning algorithm  removes a rule only if it does not correctly classify any instance from those uncovered by nonremoved rules with higher precedence. The pruning, therefore, removes only rules that do not contribute to the number of correctly classified instances, and as such, it does not increase the number of misclassified instances in the training dataset. Similarly, the default rule pruning ensures that the replacement of the tail of the rules in the base classifier by the default rule does not increase the count of misclassified instances in the training dataset.

\subsection{Default Rule Overlap Pruning }
The default rule overlap checks whether when some of the rules  classifying into the same class as the default rule  are removed the classifications would not change since the presence of the default rule would ensure that the instances are correctly classified. A \emph{pruning candidate} -- a rule that is a candidate for removal -- can be removed only if the removal does not change the classification of  instances that  are correctly classified by the pruning candidate by rules with a lower precedence.
We consider two versions of default rule overlap pruning: instance-based and range-based.

 \begin{figure}[h!]
         \centering
     \includegraphics[width=\textwidth]{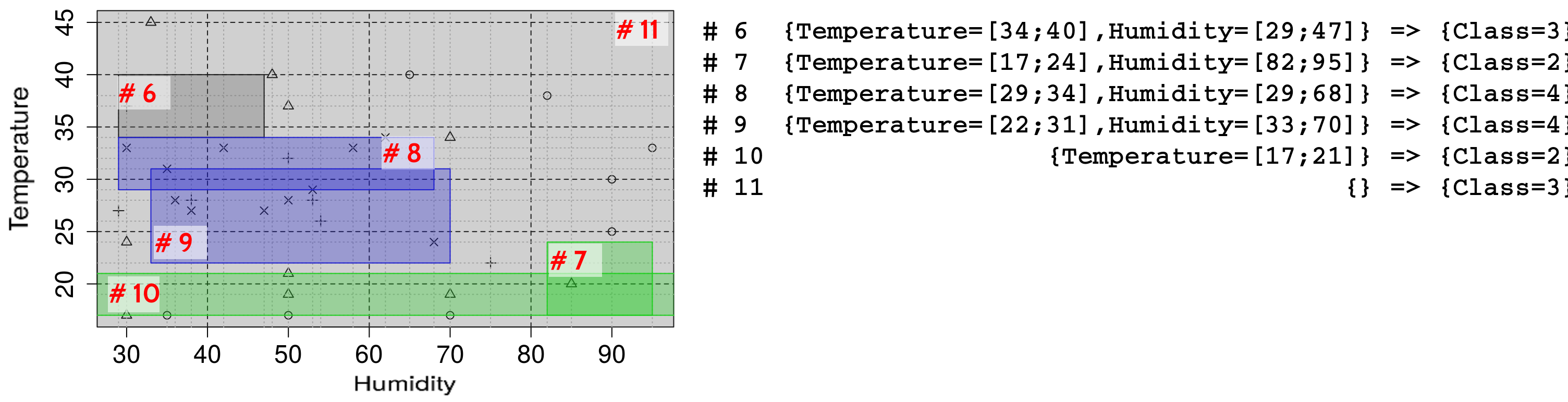}
     \caption{Illustration of default rule overlap pruning. This figure uses  a different rule list than the previous examples. The rule list is shown from rule \#6 (previous rules are not relevant). }
     \label{fig:default}
 \end{figure}

The \emph{instance-based} version, described in Procedure~\ref{alg:defRuleOverlapTrans}, removes a rule if there is no instance in the \emph{training data} that would be misclassified as a result of the removal. 

\paragraph{Example}
Referring to Fig.~\ref{fig:default}, Rule \#6 is the only pruning candidate since other rules classify to different classes than the default rule (Rule \#11). Because none of the rules between \#6 and \#11 would cause misclassification of training instances covered by \#6 if \#6 is removed, Rule \#6 is removed by instance-based pruning.

\begin{algorithm}
\caption{Default Rule Overlap Pruning (Instance-based) \emph{drop-in()}}
\label{alg:defRuleOverlapTrans}
\begin{algorithmic}[1]
\REQUIRE \emph{rules} to be pruned, $T$ training instances
\ENSURE pruned $rules$ (some elements of $rules$ removed)
\STATEx 
\STATE  $r_d \leftarrow$ default rule in $rules$  \Comment{Rule with the lowest precedence which has empty antecedent}
\FORALL{$r \in  rules \setminus r_d : cons(r) = cons(r_d)$}  \COMMENT{Iterate through pruning candidates from the highest precedence to lowest}
    \STATE $T_{r}^{corr} \leftarrow \{o\in T: o \mbox{ correctly classified by } r\}$ \label{l:trc}
    \STATE $ R_{clash}\leftarrow \{r' \in rules: r'  \mbox{  has a lower precedence than } r  \mbox{  and }  cons(r') \neq cons(r_d)\}$ \Comment{$R_{clash}$ contains candidate clashes -- rules that may misclassify instances correctly classified by $r$ if $r$ is removed}
    \STATE  $T_{clash} \leftarrow \{o \in T: o \mbox{ covered by } r' \in  R_{clash}$\} 
    \IF {$T_{clash} \cap T_{r}^{corr} = \emptyset$  } 
    \STATE $rules$ $\leftarrow$ $rules \setminus \{ r \}$ \COMMENT{No  instance correctly  classified by \emph{r} in T is misclassified if \emph{r} is removed.} 
    \ENDIF

    \ENDFOR   
 \STATE  \textbf{return}  \emph{rules}
\end{algorithmic}
\end{algorithm}

The \emph{range-based} version, depicted in Procedure~\ref{alg:defRuleOverlapRange} (located in the Appendix), analyses overlaps in the range of literals between the \emph{pruning candidate}  and all  \emph{potentially clashing rules}. A potential clashing rule is a rule with lower precedence than the pruning candidate with respect to the criteria in Definition~\ref{def:ranking} that has a different class in the consequent than the pruning candidate.  The pruning candidate is removed  only if none of the potential clashing rules overlaps the region covered by the pruning candidate. 

Range-based pruning imposes more stringent conditions for the removal of a rule than instance-based pruning as it checks empty overlap in regions covered by the rule (as opposed to overlap in training instances). 
As  seen in our evaluation (Section~\ref{sec:experiments}), as a result of these stricter conditions it  prunes many fewer rules.	

A discussion of the limitations of the presented default rule pruning algorithms is provided in Section~\ref{sec:limitations}.

\paragraph{Example}
Referring to data in Fig.~\ref{fig:default} left, rules between \#6 and \#11 seem to cover different geometric regions, which would be an argument for removing \#6. After closer inspection (Fig.~\ref{fig:default} right) we note that \#6 overlaps in Humidity and shares an inclusive boundary on Temperature with Rule \#8, which also includes Temperature=34  but classifies to a different class. Thus, Rule \#8 is confirmed as a clashing rule for Rule \#6, and since a clashing rule exists, \#6 is not removed using range-based pruning.

\emph{Effect on measures of classifier quality}
In the \emph{range-based} version, only rules that cover redundant regions (in terms of literal ranges) are removed; thus, pruning does not have an effect on the accuracy of the pruned rule list on both training and unseen  data.  The more relaxed 
\emph{instance-based} version determines rule redundancy based on the training instances, which only results in no change in the accuracy on the training data.

\section{Experiments}
\label{sec:experiments}
In this section, we present an evaluation of the presented rule tuning steps comprising the QCBA framework.
The scope of the evaluation, in terms of the number and character of the included datasets  and reference baselines, follows the approach taken in related research. 

In \cite{lakkarajuinterpretable}, the results of IDS were compared against seven other algorithms. Of these, five were general machine learning algorithms (such as random forests, and decision trees), and two were closely related -- BDL \cite{letham2015interpretable} and CBA.
In \cite{yang2017scalable}, the results of SBRL are compared against nine other algorithms, out of these seven were general machine learning algorithms (such as C4.5, and RIPPER), and two were closely related (CBA and CMAR).

We perform two types of benchmarks. The first type executes the baseline algorithm and consequently postprocessed the resulting models by QCBA comparing the results. In this way, we benchmark CBA, two closely related recent ARC algorithms (IDS and SBRL), two other  ARC algorithms (CPAR, CMAR), and two related inductive rule learning algorithms (FOIL2 \cite{quinlan1993foil} and FOIL enhancement PRM \cite{yin2003cpar}).
In the second type of benchmark, we do not perform the postprocessing with QCBA. This covers four standard symbolic -- intrinsically explainable -- learning algorithms (FURIA, RIPPER, PART, C4.5/J48) as well as a state-of-the-art rule learner CORELS.

As for the number of datasets, IDS  was evaluated on three proprietary datasets \cite{lakkarajuinterpretable}, SBRL on seven publicly available datasets \cite{yang2017scalable}, and CORELS   on three datasets \cite{angelino2017learning}. In our approach, we used 23 open datasets (some algorithms are evaluated only on a subset of these).

\vspace{-2mm}\subsection{Datasets and Setup}
\label{sec:uci}
The University of California provides a collection of publicly available datasets that are commonly used for benchmarking machine learning algorithms at \url{https://archive.ics.uci.edu}.  We choose 22 datasets to perform the main evaluation. The selection criteria were a) at least one numerical predictor attribute, and b) the dataset being previously used in the evaluation of symbolic learning algorithms in one of the following seminal papers: \cite{alcala2011fuzzy,huhn2009furia,Liu98integratingclassification,quinlan1996improved} (ordered by publication date).

Details of the 22 datasets selected for the main benchmark are given in Table~\ref{tbl:datasets}, where \emph{att.} denotes the number of attributes, \emph{inst.} denotes a number of instances, \emph{miss.} whether the dataset contains missing observations. As follows from the table, several datasets come from visual information processing or signal processing domains (ionosphere, letter, segment, sonar). The second most strongly represented is the medical domain (colic, breast-w, diabetes, heart-statlog, lymph). Eleven datasets are binary classification problems, nine datasets are multinominal (more than two classes), and two datasets have an ordinal class attribute (autos and labour). 

In addition, the 23rd dataset (intrusion detection) is used for the scalability analysis in Section~\ref{ss:scalability}.
\begin{table}[h!]
{\footnotesize
\begin{center}
\begin{tabular}{lrrclp{4.3cm}}
\toprule
dataset & att. &  inst. & miss. & class & description \\
\midrule
anneal & 39 & 898 & Y & nominal (6) & steel annealing dataset \\ 
australian & 15 & 690 & N & binary &credit card applications\\ 
autos & 26 & 205 & Y & ordinal (7) &riskiness of second hand cars \\ 
breast-w & 10 & 699 & Y & binary & breast cancer\\ 
colic & 23 & 368 & Y & binary & horse colic (surgical or not)\\ 
credit-a & 16 & 690 & Y & binary & credit approval \\ 
credit-g & 21 & 1000 & N & binary & credit risk \\ 
diabetes & 9 & 768 & N & binary & diabetes \\ 
glass & 10 & 214 & N& nominal (6) & types of glass \\ 
heart-statlog & 14 & 270 & N & binary & diagnosis of heart disease \\ 
hepatitis & 20 & 155 & Y & binary & hepatitis prognosis (die/live)\\ 
hypothyroid & 30 & 3772 & Y & nominal (3) & thyroid disease data set \\ 
ionosphere & 35 & 351 & N & binary & radar data\\ 
iris & 5 & 150 & N & nominal (3) &iris (flower) varieties\\ 
labor & 17 & 57 & Y & ordinal (3) &  contributions to health plan \\ 
letter & 17 & 20000 & N  & nominal (26) & letter recognition\\ 
lymph & 19 & 148 & N & nominal (4)& lymphography domain\\ 
segment & 20 & 2310 & N & nominal (7) &  image segment classification \\ 
sonar & 61 & 208 & N & binary & object based on sonar signal \\ 
spambase & 58 & 4601 & N& binary & spam detection\\ 
vehicle & 19 & 846 & N & nominal (4) & object type based on silhouette \\ 
vowel & 13 & 990 & N & nominal (11) & vowel recognition \\ 
\bottomrule

\end{tabular}
\end{center}
}
\caption[Overview of 22 datasets involved in the benchmark]{Overview of 22 datasets involved in the main benchmark. }
\label{tbl:datasets}
\end{table}

Numerical (quantitative) explanatory attributes  with three or more distinct values are subject to discretisation using the MDLP algorithm \cite{conf/ijcai/FayyadI93}. We use the MDLP implementation wrapped in our \texttt{arc} package. Prediscretised data is used only for association rule classification algorithms (CBA, IDS, and SBRL).  The remaining algorithms involved in the benchmark do not require prediscretisation. 
The evaluation is performed using  10-fold stratified cross-validation. All evaluations used the same folds. 

The results are obtained using the open-source CBA and QCBA implementations in our \texttt{arc} and \texttt{qCBA} packages, which we made available in the CRAN repository of R language. The core of QCBA is implemented in Java 8.  For CMAR, CPAR, PRM and FOIL we used \texttt{arulesCBA} package \cite{arulesCBA}, which is based on the LUCS-KDD Java implementation of these algorithms.\footnote{\url{https://cgi.csc.liv.ac.uk/~frans/KDD/Software/}}

The evaluations were performed on a computer running Ubuntu 20.04 equipped  with 32 GB of RAM, a Samsung SSD 960 EVO hard disk,  and an Intel(R) Core(TM) i5-7440HQ CPU @ 2.80GHz. For the evaluations in Table~\ref{tbl:qcba} and Table~\ref{tbl:olderBench} the software was externally restricted to one CPU core. The scalability benchmark (Figures~\ref{fig:indivallKDDablation},\ref{fig:kdd}) was run on an Intel(R) Xeon(R) CPU E5-2630 v4 @ 2.20GHz with six virtualized cores and 22 GBs of RAM.  
\paragraph{Metaparameter Setup}
The CBA algorithm has three main hyperparameters -- minimum confidence, minimum support thresholds, and the total number of candidate rules. In \cite{Liu98integratingclassification} it was recommended to use 50\% as the minimum confidence and 1\% as the minimum support. For our experiments, we used these thresholds.  In \cite{Liu98integratingclassification},  the total number of rules used was 80.000; however, it was noted that the performance starts to stabilise at approximately 60.000 rules. According to our experiments (not included in this article), there is virtually no difference between the 80.000 and 50.000 thresholds apart from the higher computation time for the former; therefore we used 50.000.\footnote{In our best setup, we observed less than 0.1 percent point improvement in the average accuracy and 5 percent point increase in the average rule count when the maximum number of rules is increased from 50.000 to 80.000.} We also limited the maximum number of literals in the antecedent to 5. For the standalone CBA run, default rule pruning is used. For the other runs, it is not performed within CBA.\footnote{Preferably,  QCBA should obtain its input model built with CBA but with default rule pruning not performed as part of CBA to obtain more rules for tuning. Such variation of CBA is not separately described in \cite{Liu98integratingclassification} or in other prior research known to us.
Our CBA implementation is adapted to allow the deactivation of default rule pruning. }

\subsection{Effects of individual steps}
\label{sec:marcevalsteps}

The proposed tuning steps do not have any mandatory thresholds. The extension process contains two numerical parameters, which are left set to their default values: \emph{minImprovement=0} and \emph{minCI=-1}. These default values are justified in Section~\ref{ss:Ext}. The effect of varying  the latter threshold is studied in Subs.~\ref{ss:scalability}. 

\paragraph{Evaluation Methodology}
We evaluate the individual rule tuning steps. As a baseline, we use a CBA run with default parameters corresponding to those recommended in \cite{Liu98integratingclassification} (with small deviations justified above). 
Classification performance is measured by accuracy as in several studies of ARC classifier performance, such as \cite{yang2017scalable,angelino2017learning,wang2018multi}. 
All results are reported using tenfold cross-validation with macro averaging. All evaluations used the same folds.  The average accuracy for all 22 datasets is reported as an indicative comparison measure. We also include a won-tie-loss matrix, which compares two classifiers by reporting the number of datasets where the reference classifier wins, loses,  or the two classifiers perform equally well. 
For a more reliable comparison, we include  p-value for the Wilcoxon signed-rank test computed on accuracies. This test is preferred over Friedman's test \cite{benavoli2016should}.

We use three metrics to measure the size of the model: the average antecedent length (number of conditions in the rule),  the average number of rules per model and the average number of conditions per model, which is computed as the \emph{average number of rules} $\times$ \emph{average antecedent length}.  These are the most common measures in recent related research: \cite{angelino2017learning,lakkarajuinterpretable,wang2018multi,lakkarajuinterpretable}.

We also include a benchmark indicating the computational requirements of the individual proposed postprocessing steps. The learning times reported in Table~\ref{tbl:qcba} are computed as an average of classifier learning time for 220 models (10 folds for each of the 22 datasets).  In addition to the absolute run time, which can be volatile across software and hardware platforms, we include the relative execution time with the CBA baseline that is assigned a score of 1.0.

The CBA baseline includes the discovery of candidate association rules, data coverage pruning, and default rule pruning.  
It should be noted that the implementation of the tuning steps is not heavily optimised for speed.

\paragraph{Overview of the results}
As a first step in determining the effect of the individual tuning steps, we choose which ARC algorithm would be used to generate base models that would be postprocessed. We chose CBA, which is the most commonly used reference algorithm in related research and is still considered state-of-the-art \cite{lakkarajuinterpretable}.

\setlength{\tabcolsep}{4pt}
\begin{table}[h!]
\centering

{\footnotesize
\begin{tabular}{lcccccccc}
\toprule
configuration & cba & \multicolumn{1}{c}{\#1} & \multicolumn{1}{c}{\#2} & \multicolumn{1}{c}{\#3} & \multicolumn{1}{c}{\#4} & \multicolumn{1}{c}{\#5} & \multicolumn{1}{c}{\#6} & \multicolumn{1}{c}{\#7} \\ 
\midrule
\multicolumn{8}{c}{CBA as a baseline and a source of input candidate rules for postprocessing} \\
\midrule
data coverage pruning & Y & Y & Y & Y & Y & Y & Y & Y \\
default rule pruning  & Y & - & - & - & - & - & - & - \\
\midrule
\multicolumn{8}{c}{QCBA postprocessing steps} \\
\midrule
refit & na & Y & Y & Y & Y & Y & Y & Y \\ 
literal pruning& na & - & Y & Y & Y & Y & Y & Y \\ 
trimming & na & - & - & Y & Y & Y & Y & Y \\ 
extension & na & - & - & - & Y & Y & Y & Y \\ 
post-pruning -- data coverage  & na & - & - & - & - & Y & Y & Y \\
post-pruning -- default rule  & na & - & - & - & - & Y & Y & Y \\
def. rule overlap --  instance & na & - & - & - & - & - & Y & - \\ 
def. rule overlap -- range & na  & - & - & - & - & - & - & Y \\ 
\midrule
\multicolumn{8}{c}{Results} \\
\midrule
won/tie/loss (QCBA vs base) & base &9/6/7 & 9/6/7 & 7/7/8 & 8/7/7 & 10/9/3 & 7/6/9 & 10/9/3 \\
p-value (QCBA vs base) & base & \multicolumn{1}{r}{0.58} & \multicolumn{1}{r}{0.58} & \multicolumn{1}{r}{0.62} & \multicolumn{1}{r}{0.82} & \multicolumn{1}{r}{0.23} & \multicolumn{1}{r}{0.09} & \multicolumn{1}{r}{0.23} \\
accuracy & \multicolumn{1}{r}{0.81} & \multicolumn{1}{r}{0.81} & \multicolumn{1}{r}{0.81} & \multicolumn{1}{r}{0.81} & \multicolumn{1}{r}{0.81} & \multicolumn{1}{r}{0.81} & \multicolumn{1}{r}{0.80} & \multicolumn{1}{r}{0.81} \\
avg AUC for binary class datasets & \multicolumn{1}{r}{0.82} & \multicolumn{1}{r}{0.82} & \multicolumn{1}{r}{0.82} & \multicolumn{1}{r}{0.83} & \multicolumn{1}{r}{0.82} & \multicolumn{1}{r}{0.82} & \multicolumn{1}{r}{0.80} & \multicolumn{1}{r}{0.82} \\
avg conditions / rule & \multicolumn{1}{r}{2.8} & \multicolumn{1}{r}{2.8} & \multicolumn{1}{r}{2.8} & \multicolumn{1}{r}{2.8} & \multicolumn{1}{r}{2.8} & \multicolumn{1}{r}{2.8} & \multicolumn{1}{r}{2.8} & \multicolumn{1}{r}{2.8} \\
avg number of rules & \multicolumn{1}{r}{84.2} & \multicolumn{1}{r}{91.8} & \multicolumn{1}{r}{91.8} & \multicolumn{1}{r}{91.8} & \multicolumn{1}{r}{91.8} & \multicolumn{1}{r}{65.9} & \multicolumn{1}{r}{47.8} & \multicolumn{1}{r}{65.5} \\
avg conditions / model & \multicolumn{1}{r}{235.3} & \multicolumn{1}{r}{258.7} & \multicolumn{1}{r}{258.7} & \multicolumn{1}{r}{258.7} & \multicolumn{1}{r}{258.7} & \multicolumn{1}{r}{183.6} & \multicolumn{1}{r}{132.0} & \multicolumn{1}{r}{182.4} \\

Median learning time [s]  & \multicolumn{1}{r}{6.2} & \multicolumn{1}{r}{0.2} & \multicolumn{1}{r}{0.2} & \multicolumn{1}{r}{0.3} & \multicolumn{1}{r}{4.7} & \multicolumn{1}{r}{4.8} & \multicolumn{1}{r}{5.0} & \multicolumn{1}{r}{5.2} \\
-- Normalized & \multicolumn{1}{r}{1.00} & \multicolumn{1}{r}{0.02} & \multicolumn{1}{r}{0.04} & \multicolumn{1}{r}{0.05} & \multicolumn{1}{r}{0.76} & \multicolumn{1}{r}{0.77} & \multicolumn{1}{r}{0.81} & \multicolumn{1}{r}{0.83} \\
\bottomrule
\end{tabular}
}
\caption{Effect of individual steps -- aggregated results for 22 UCI datasets. The reported learning times for \#1-\#7 exclude the time required to learn the input CBA rule model.  }
\label{tbl:qcba}
\end{table}

Table~\ref{tbl:qcba}  demonstrates the effect of the individual tuning steps comprising our postprocessing framework when applied to models learned with CBA. The performance is compared with CBA as a baseline. 

Configuration \#1 corresponds to the refit tuning step that is performed on top of CBA, configuration \#2 corresponds to the refit tuning step and the literal pruning, etc. Configurations \#6 and \#7 correspond to all proposed tuning steps performed (\#6 uses instance-based pruning and \#7 uses range-based default rule overlap pruning). 

The QCBA setup that  achieves the maximum reduction in the size of the classifier is configuration \#6, which  includes all the tuning steps. A compromise between accuracy and size is provided by \#5, which compared to \#6 excludes the default overlap pruning.
As follows from a comparison of \#5 and \#7, range-based pruning is ineffective  (for CBA as a base learner) on this collection of datasets. 

The literal pruning step appears redundant when CBA models are postprocessed.
However, for the postprocessing of the models generated by the IDS algorithm, this processing step has a positive effect as further elaborated in  Section~\ref{ss:ids}.

\emph{Accuracy vs. AUC}
We experimented with two ways of computing the confidence of classification, which is the basis for computing the AUC values.\footnote{One option was to use the ``global'' (established on the entire training set) confidence of the firing rule, as the confidence of the classification. In the second ``ordered confidence'' option (shown in the table), the confidence of the classification was computed as the confidence of the firing rule, which was computed on the set of instances in the training data reaching this rule. The difference between the ordered and global AUC was less than 0.01. More details on the computation are in the documentation of the supplementary R packages implementing CBA and QCBA. }
In both cases, the comparison of AUC values had a similar outcome as a comparison based on accuracy.
\paragraph{Model size vs. predictive performance}
The full QCBA postprocessing (\#6) reduces the size of the model with an average of 235.3 conditions  by 44\% compared to  the original CBA model to 132.0 conditions,  incurring a point drop of only a 1\% in the average accuracy.
QCBA configuration \#5 results in a reduction  in size of 22\%. Additionally,  configuration \#5 has a better accuracy on 10 datasets while CBA wins only on 3 datasets, leaving 9 results in a draw.

The reduction in model size is due to the removal of rules. The average rule length of the postprocessed CBA models remains the same. However, as we show later when models produced by IDS and SBRL are postprocessed, the  average rule length also drops.  

It should be noted that the CBA baseline is generated with default rule pruning, while postprocessing with QCBA default rule pruning within CBA is disabled since QCBA performs default rule pruning as its last pruning step. As a result, the number of rules in QCBA configurations \#1 to \#4 increase when compared to CBA. 

\paragraph{Runtime}
\label{ss:runtimebench}
The results for the runtime are reported in the last two rows of Table~\ref{tbl:qcba}. 
It can be seen that refit, literal pruning and trimming take together less time than learning a CBA model alone. 
The most computationally intensive operation is the extension. If we look at the median learning times, we can observe that the postprocessing takes about as long as it takes to learn the input CBA model.

 \begin{figure}
         \centering
     \includegraphics[width=\textwidth]{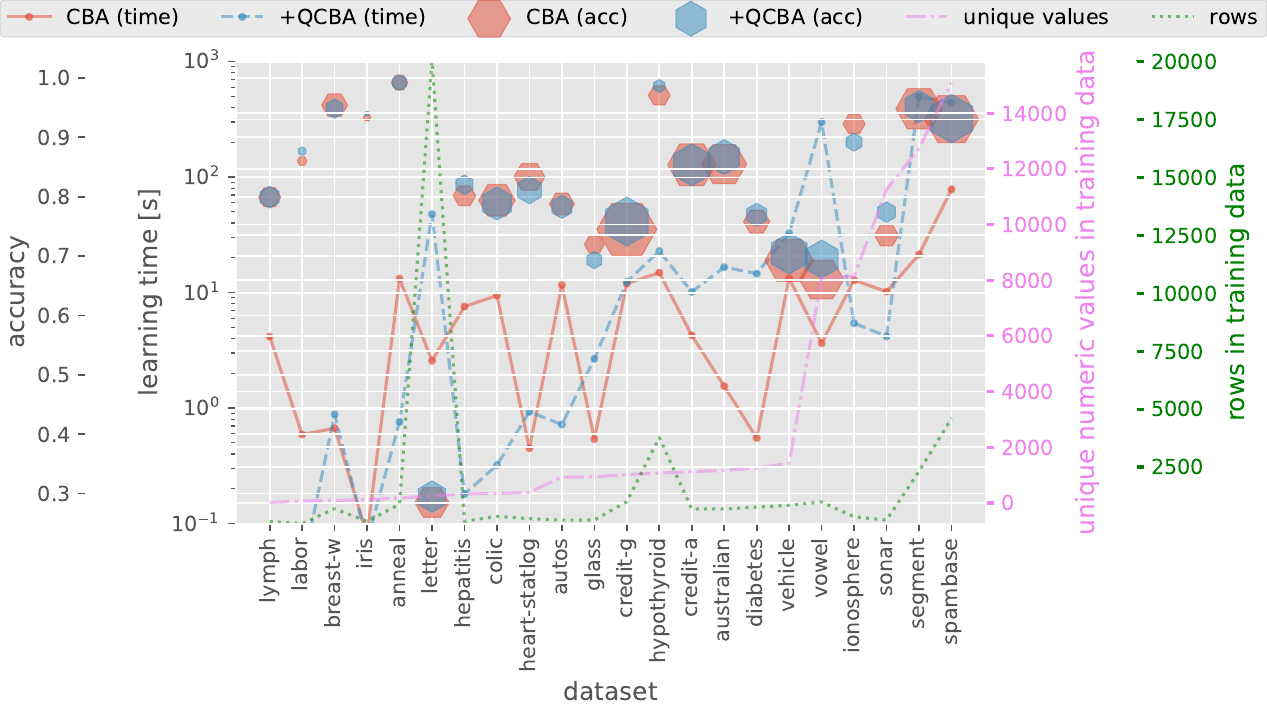}
     \caption{CBA+QCBA\#5 on individual datasets.}
     \label{fig:indivall}
 \end{figure}
 
Figure~\ref{fig:indivall} shows the breakdown results for individual datasets. In this figure, the size of the hexagonal symbol denotes the size of the model, computed as the number of rules $\times$ average number of conditions (smaller is better). The QCBA learning time reported is for postprocessing only. The purple line denotes the number of unique values across all attributes (except the target) in the given dataset and the  datasets (on the X-axis) are sorted by this value.

The larger run time for several datasets in Figure~\ref{fig:indivall} can be attributed to the extension step. The graph also shows  the largest impact on datasets with many distinct values or rows, which correspond to segment, letter and spambase datasets. The segment and letter datasets contain various image metrics and spambase word frequency attributes.  Such datasets are not typical representatives of the use cases, where interpretable machine learning models are needed. Nevertheless, the evaluation of the runtime indicates that the computational optimisation of the extension algorithm is one of the most important areas for further work.

\vspace{-2mm}\subsection{Postprocessing SBRL Models}
\label{ss:sbrl}
The Scalable Bayesian Rule Lists (SBRL) \cite{yang2017scalable} is a recently proposed rule learning algorithm. 
As with most association rule learning approaches, the SBRL  algorithm can process only nominal attributes.
In this experiment, we postprocess models generated by the SBRL with the proposed rule tuning steps. 

Since SBRL is limited to datasets with binary classes, we process all eleven datasets with binary class labels from Table~\ref{tbl:datasets}. R Implementation \texttt{sbrl} (CRAN version 1.2) from the SBRL authors was used.\footnote{\url{https://cran.r-project.org/web/packages/sbrl/}}
The postprocessed models are generated using the QCBA package referenced earlier. For prediction, rules are applied in the order output by QCBA.
We evaluated the two best QCBA configurations from the CBA evaluation (QCBA \#5 and QCBA \#6 from Table~\ref{tbl:qcba}).

\paragraph{Metaparameter setup}
SBRL is run with the following default parameter settings: 
{\footnotesize
\texttt{iters=30000, pos\_sign=0, neg\_sign=1, rule\_minlen=1, rule\_maxlen=1,    eta=1.0,                      minsupport\_pos=0.10, lambda=10.0,\\ minsupport\_neg=0.10,    alpha=\{1,1\}, nchain=10.}}
With this setting, which limits the antecedent length to 1, the SBRL produces very simple models with a somewhat lower accuracy. The second evaluated setting for the SBRL differs only by the increased maximum rule length, which allows the algorithm to learn more expressive models.  The \texttt{rule\_maxlen} parameter is set to 10 for all datasets, except those where this setting resulted in an out of memory error in the first rule generation phase within the SBRL.\footnote{For hepatitis, the threshold was decreased to 5, for ionosphere and sonar datasets the threshold was decreased to 3, and for spambase it was decreased to 2.}

\paragraph{Results}
\label{par:sbrl}
Table \ref{tbl:sbrl} shows the predictive performance and comprehensibility metrics for the SBRL only, the SBRL model postprocessed with all the proposed tuning steps except default rule overlap pruning (SBRL +QCBA\#5), and the SBRL model postprocessed with all proposed tuning steps including instance-based default rule overlap pruning (SBRL+QCBA\#6). The table shows results for two SBRL configurations: rules of length 1, rules of length 10 (with several exceptions noted above). The learning time for the QCBA runs excludes the time required to build the input SBRL model. The p-value is for Wilcoxon signed-rank test computed on accuracies. The won/tie/loss record denotes the number of times that the postprocessed model has a better, same, or lower accuracy than the input SBRL model.  The model size is computed as a product of the average rule count and the average rule length. The learning times for postprocessing do not include the time required to build the input SBRL model. The normalised learning time is reported relative to the learning time of the SBRL only.

\begin{table}[h!]
\centering

\begin{tabular}{lp{1cm}p{1.2cm}p{1.2cm}|p{1.2cm}p{1.2cm}p{1.2cm}}
\toprule
 & \multicolumn{3}{c}{SBRL trained on short rules} & \multicolumn{3}{c}{SBRL trained on long rules} \\  
 \midrule
 & only SBRL & \multicolumn{1}{p{1cm}}{QCBA (+Q\#5)} & \multicolumn{1}{p{1cm}}{QCBA (+Q\#6)} & only SBRL & \multicolumn{1}{p{1cm}}{QCBA (+Q\#5)} & \multicolumn{1}{p{1cm}}{QCBA (+Q\#6)} \\ 
 \midrule
 accuracy & \multicolumn{1}{r}{0.80} & \multicolumn{1}{r}{0.81} & \multicolumn{1}{r}{0.79} & \multicolumn{1}{r}{0.81} & \multicolumn{1}{r}{0.81} & \multicolumn{1}{r}{0.80} \\ 
won/tie/loss (base+QCBA vs base) & \multicolumn{1}{r}{base } & \multicolumn{1}{r}{6-5-0} & \multicolumn{1}{r}{2-5-4} & \multicolumn{1}{r}{base} & \multicolumn{1}{r}{7-1-3} & \multicolumn{1}{r}{6-1-4} \\ 
p-value (base+QCBA vs base)& \multicolumn{1}{r}{base} & \multicolumn{1}{r}{0.03} & \multicolumn{1}{r}{0.25} & \multicolumn{1}{r}{base} & \multicolumn{1}{r}{0.75} & \multicolumn{1}{r}{0.72} \\ 
avg number of rules & \multicolumn{1}{r}{4.8} & \multicolumn{1}{r}{3.7} & \multicolumn{1}{r}{3.4} & \multicolumn{1}{r}{3.3} & \multicolumn{1}{r}{3} & \multicolumn{1}{r}{2.7} \\ 
avg conditions / rule & \multicolumn{1}{r}{0.8} & \multicolumn{1}{r}{0.7} & \multicolumn{1}{r}{0.6} & \multicolumn{1}{r}{1.5} & \multicolumn{1}{r}{1.2} & \multicolumn{1}{r}{1.2} \\ 
avg conditions / model & \multicolumn{1}{r}{3.7} & \multicolumn{1}{r}{2.5} & \multicolumn{1}{r}{2.2} & \multicolumn{1}{r}{4.8} & \multicolumn{1}{r}{3.6} & \multicolumn{1}{r}{3.2} \\
median learning time [s] & \multicolumn{1}{r}{0.6} & \multicolumn{1}{r}{0.1} & \multicolumn{1}{r}{0.1} & \multicolumn{1}{r}{24.9} & \multicolumn{1}{r}{0.1} & \multicolumn{1}{r}{0.1} \\ 
 - normalized & \multicolumn{1}{r}{1.00} & \multicolumn{1}{r}{0.17} & \multicolumn{1}{r}{0.17} & \multicolumn{1}{r}{1.00} & \multicolumn{1}{r}{0} & \multicolumn{1}{r}{0} \\ 

\bottomrule
\end{tabular}
\caption{Postprocessing Scalable Bayesian rule list (SBRL) models. Aggregated results for 11 binary-class datasets from Table~\ref{tbl:datasets}.}
\label{tbl:sbrl}
\end{table}

Similarly to the study of the effect of individual steps performed for CBA, it is reported in Table~\ref{tbl:sbrl} that configuration QCBA\#6 with all tuning steps enabled generates the smallest models, while configuration \#5 with all tuning steps, but without default rule overlap pruning, provides balance between accuracy and size. 

Configuration QCBA\#6 results in a reduction in size, as represented by the average number of conditions in the models, by 33\% to 41\%, while configuration \#5 results in a reduction by 25\% to 32\% but has higher gains in predictive performance.  According to the  won/tie/loss record in Table~\ref{tbl:sbrl},   the SBRL models postprocessed by the proposed tuning steps have a higher accuracy on most datasets compared to the SBRL-only models. The postprocessing is most effective on models composed of short rules: there, the improvement is statistically significant ($p < 0.05$).

\begin{figure}
         \centering
       
     \includegraphics[width=\textwidth]{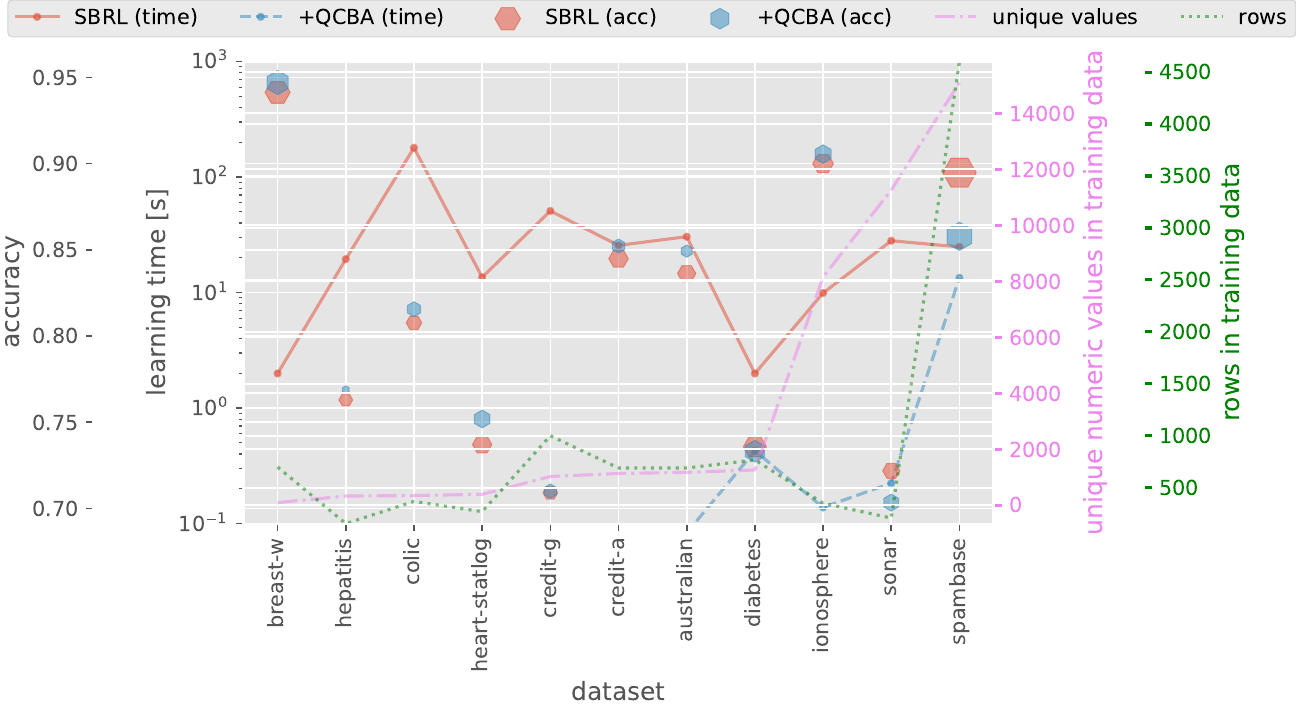}
     \caption{The SBRL (long rules)+QCBA\#5 on individual datasets. }
     \label{fig:indivallSBRL}
     
 \end{figure}

Overall, the evaluation shows that postprocessing the SBRL models with the proposed tuning steps results in a reduced model size and an improved accuracy.  The computationally most intensive part is the generation of input rules performed within the SBRL. As Table~\ref{tbl:sbrl} and Fig.~\ref{fig:indivallSBRL} show,  the additional computational cost of postprocessing the SBRL models is relatively low, even when the SBRL setup involves learning of longer input rules. The reason is that the models on the SBRL output consistently contain a small number of rules, which eases their subsequent tuning. As Figure~\ref{fig:indivallSBRL} shows, there is a dependency between the number of distinct numerical values in the dataset and the learning time.

\subsection{Postprocessing IDS models}
\label{ss:ids}
Similar to the SBRL and CBA, the Interpretable Decision Sets  (IDS) \cite{lakkarajuinterpretable} is an association rule classification algorithm.  
In the first step, candidate association rules are generated from frequent itemsets returned by the Apriori algorithm. In the rule optimisation step, a subset of the generated rules is selected to form the final classifier. The rule selection procedure is a subject of optimisation using the smooth local search  (SLS) algorithm \cite{feige2011maximizing}, which guarantees a near-optimal solution and is at least 2/5 of the optimal solution. IDS uses a compound objective function composed of several interpretability subobjectives (number of rules,  rule length, minimise rule overlap) and accuracy (maximise precision and recall).

\begin{table}[h!]
\centering

\begin{tabular}{lp{1.2cm}p{1.2cm}p{1.2cm}p{1.2cm}p{1.2cm}} \toprule
                          & \multicolumn{1}{p{1cm}}{only IDS} & \multicolumn{1}{p{1.2cm}}{QCBA (+Q\#5)}    & \multicolumn{1}{p{1.2cm}}{QCBA (+Q\#5)}                    & \multicolumn{1}{p{1.2cm}}{QCBA (+Q\#6)}               & \multicolumn{1}{p{1.2cm}}{QCBA (+Q\#6)}      
                          \\
literal pruning           & \multicolumn{1}{c}{NA}       & \multicolumn{1}{c}{-}      & \multicolumn{1}{c}{Y}      & \multicolumn{1}{c}{-}      & \multicolumn{1}{c}{Y}       \\ \midrule 
accuracy  & \multicolumn{1}{r}{0.61}                         & \multicolumn{1}{r}{0.63}                       & \multicolumn{1}{r}{0.63}                       & \multicolumn{1}{r}{0.63}                       & \multicolumn{1}{r}{0.63}                        \\
won/tie/loss (base+QCBA vs base)              &  \multicolumn{1}{r}{base}         & \multicolumn{1}{r}{14/7/1} & \multicolumn{1}{r}{13/8/1} & \multicolumn{1}{r}{13/8/1} & \multicolumn{1}{r}{12/9/1}  \\
p-value (base+QCBA vs base)                  & \multicolumn{1}{r}{base}        & \multicolumn{1}{r}{0.00}                    & \multicolumn{1}{r}{0.00}                    & \multicolumn{1}{r}{0.00}                    & \multicolumn{1}{r}{0.00}                     \\
avg number of rules       & \multicolumn{1}{r}{16.6}                         & \multicolumn{1}{r}{4.6}                        & \multicolumn{1}{r}{4.2}                        & \multicolumn{1}{r}{3.7}                        & \multicolumn{1}{r}{3.4}                         \\
avg conditions / rule     & \multicolumn{1}{r}{3.6}                          & \multicolumn{1}{r}{2.1}                        & \multicolumn{1}{r}{1.7}                        & \multicolumn{1}{r}{1.9}                        & \multicolumn{1}{r}{1.5}                         \\
avg conditions / model    & \multicolumn{1}{r}{60.4}                         & \multicolumn{1}{r}{9.6}                       & \multicolumn{1}{r}{7.1}                         & \multicolumn{1}{r}{7.2}                        & \multicolumn{1}{r}{5.3}                         \\
median learning time [s]     & \multicolumn{1}{r}{21.1}                         & \multicolumn{1}{r}{2.8}                        & \multicolumn{1}{r}{1.3}                        & \multicolumn{1}{r}{2.8}                        & \multicolumn{1}{r}{1.5}                         \\
 - normalized & \multicolumn{1}{r}{1.00}                            & \multicolumn{1}{r}{0.13}                       & \multicolumn{1}{r}{0.06}                       & \multicolumn{1}{r}{0.13}                       & \multicolumn{1}{r}{0.07}             \\          \bottomrule
\end{tabular}
\caption{Postprocessing Interpretable Decision Sets (IDS) models.Aggregate results for 22 datasets from Table~\ref{tbl:datasets}.}
\label{tbl:ids}
\end{table}

\paragraph{Setup}
For evaluation purposes, we use our reimplementation of the IDS algorithm described in \cite{idsRULEML}.\footnote{\url{https://github.com/jirifilip/pyIDS} -- version from December 9, 2020} The IDS authors have made a reference implementation available on GitHub: nevertheless the evaluation reported in \cite{idsRULEML} shows that the reference implementation is too slow to be applied on the benchmark datasets introduced in Table~\ref{tbl:datasets}.
While the implementation described in our paper \cite{idsRULEML} is faster, and there are additional performance gains after  \cite{idsRULEML} was published, there are still performance issues.
For the optimisation to finish in a reasonable time (within minutes per dataset fold), we limit  the  input rule list to 50 rules. Rules are mined with a similar setting that is used for CBA. The minimum support threshold is set to 1\%, and we also apply a minimum confidence threshold of 50\%, which improves the results. The maximum rule length is set to 5 (4 for the spambase dataset).  All seven lambda parameters, which in the IDS essentially corresponds to the accuracy-interpretability trade-off, are set to equal values (1.0).\footnote{We also experiment with setting those lambda parameters to 0, which do not directly correspond to accuracy and model size ($\lambda_3,\lambda_4,\lambda_5$). Such setting took extremely long to optimise (did not finish for a single fold of the iris dataset within 1 hour).}

For prediction on IDS models, rules are sorted by a harmonic mean of support and confidence as specified in \cite{lakkarajuinterpretable} and applied in this order; the first firing rule classifies the instance. The rule order for prediction on models postprocessed by QCBA is determined by the order output by QCBA (criteria in Def.~\ref{def:ranking}).

\paragraph{Results}
\label{para:idsresults}
The results reported in Table~\ref{tbl:ids} indicate that the proposed tuning  steps consistently reduce the size of models generated by IDS, while also improving predictive performance on the majority of datasets. Each IDS+QCBA configuration is run in two variations without literal pruning or with it (default).
On average, the accuracy improves by 2\% (absolute), and the model size is reduced by 85\% (for QCBA\#5). Literal pruning makes a modest contribution to this reduction. Models built without literal pruning are 35\% (QCBA\#5) and 36\% (QCBA\#6) larger than models built with literal pruning (computed from average conditions/model).

The Wilcoxon signed-rank test shows a statistically significant improvement over IDS at $p<0.01$. The median of time required for postprocessing with QCBA is relatively small, as it takes about one-tenth of the time needed to learn the IDS models.

 \begin{figure}
         \centering
     \includegraphics[width=\textwidth]{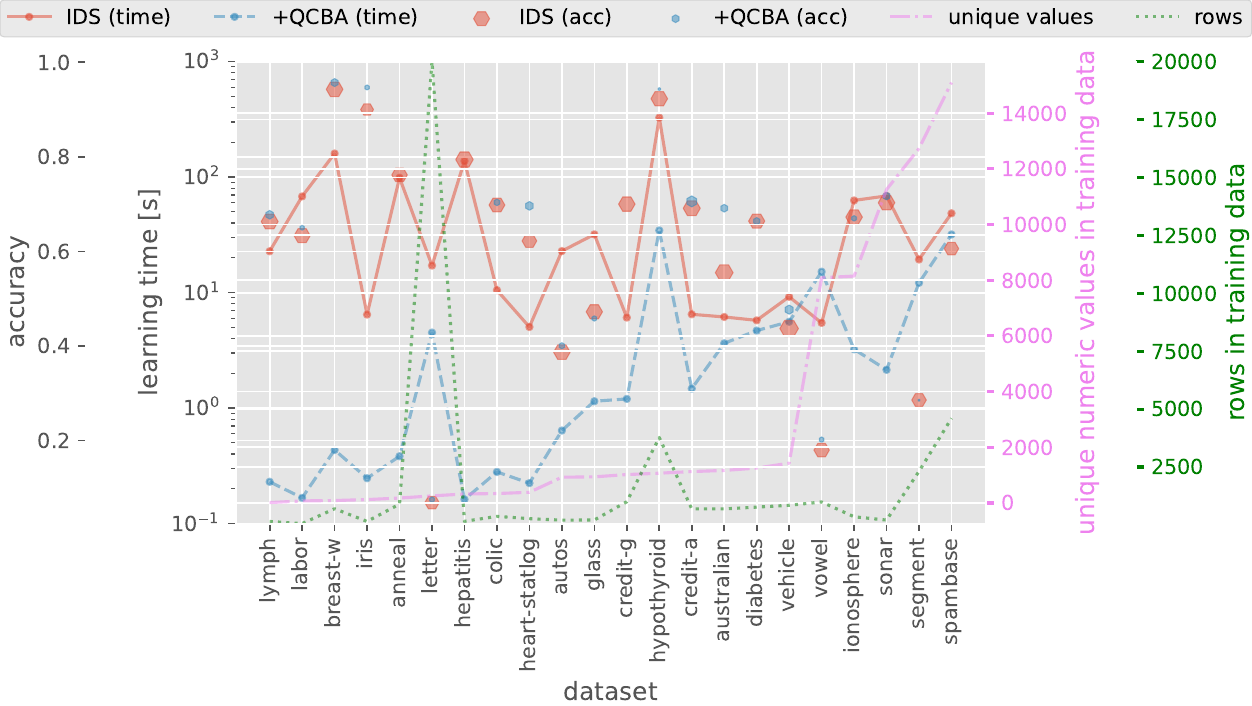}
     \caption{IDS+QCBA\#5 on individual datasets. }
     \label{fig:indivallIDS}
 \end{figure}
 
An interesting observation, which is not apparent from the results for individual datasets in Figure~\ref{fig:indivallIDS}, is that for four datasets, the QCBA postprocessing generated rule lists that are composed of only the default rule (in all folds). This empty rule has equally good accuracy as the IDS-generated rule list.\footnote{We initially suspected this is a bug in the IDS. However,  since the SLS algorithm used in IDS guarantees that the solution is at least 2/5 of the optimal solution, such a solution can still be valid.}

\vspace{-2mm}\subsection{Postprocessing CPAR, CMAR, FOIL2 and PRM models}
In this section, we evaluate against four reference algorithms widely established in the domain of rule learning.
Note that some of the baseline algorithms, such as CPAR, combine multiple rules for prediction. As part of the postprocessing, the models were converted to a QCBA model (sorted according to the CBA criteria in Def.~\ref{def:ranking}), and the prediction was based on the first rule in this list that matched the test instance.

\emph{Setup} For CMAR, CPAR, PRM and FOIL, the default metaparameter values from \cite{arulesCBA} were used. 

\emph{Results}
Table~\ref{tbl:olderBench} presents the results of postprocessing for the four rule learning algorithms. 
The results show that the QCBA reduces the average rule count for all four algorithms and the average rule length for CMAR and FOIL. The  reductions for the aggregate model size (average number of conditions in the model) range from 22\% for PRM to 79\% for CMAR. For CPAR, FOIL2 and PRM, the postprocessed models had better accuracy and won-tie-loss score. For CPAR, FOIL2 and PRM the difference is statistically significant $(p<0.05)$.  The best overall result in terms of accuracy and won-tie-loss record across all evaluated methods is obtained by FOIL2 postprocessed by QCBA.

\begin{table}[]

\begin{tabular}{p{4.6cm}p{0.8cm}p{1cm}|p{0.8cm}p{1cm}|p{0.8cm}p{1cm}|p{0.8cm}p{1cm}}
\toprule
 
                              & only CMAR   &  +Q\#5 & only CPAR  &  +Q\#5 & only FOIL2 &  +Q\#5 & only PRM   &  +Q\#5 \\ \midrule
    accuracy      & \multicolumn{1}{r}{0.83}   & \multicolumn{1}{r|}{0.83}          & \multicolumn{1}{r}{0.81}  & \multicolumn{1}{r|}{0.83}          & \multicolumn{1}{r}{0.82}  & \multicolumn{1}{r|}{0.84}           & \multicolumn{1}{r}{0.81}  & \multicolumn{1}{r}{0.83}         \\
won/tie/loss  (base+QCBA vs base)                 &\multicolumn{1}{r}{base}& \multicolumn{1}{r|}{10/2/10}       &\multicolumn{1}{r}{base}& \multicolumn{1}{r|}{15/5/2}        &\multicolumn{1}{r}{base}& \multicolumn{1}{r|}{14/5/3}         &\multicolumn{1}{r}{base}& \multicolumn{1}{r}{14/4/4}       \\
p-value (base+QCBA vs base) &\multicolumn{1}{r}{base       }& \multicolumn{1}{r|}{0.94}       &\multicolumn{1}{r}{base}& \multicolumn{1}{r|}{0.00}       &\multicolumn{1}{r}{base}& \multicolumn{1}{r|}{0.02}        &\multicolumn{1}{r}{base}& \multicolumn{1}{r}{0.01}      \\
avg number of rules           & \multicolumn{1}{r}{489.2}  & \multicolumn{1}{r|}{112.7}         & \multicolumn{1}{r}{88.5}  & \multicolumn{1}{r|}{61.9}          & \multicolumn{1}{r}{107.1} & \multicolumn{1}{r|}{76.9}           & \multicolumn{1}{r}{80.1} & \multicolumn{1}{r}{61.5}         \\
avg conditions / rule         & \multicolumn{1}{r}{3.0}      & \multicolumn{1}{r|}{2.7}           & \multicolumn{1}{r}{2.0}     & \multicolumn{1}{r|}{2.0}             & \multicolumn{1}{r}{2.5}   & \multicolumn{1}{r|}{2.3}            & \multicolumn{1}{r}{2.0}   & \multicolumn{1}{r}{2.0}            \\
avg conditions / model  & \multicolumn{1}{r}{1462.0} & \multicolumn{1}{r|}{302.6} & \multicolumn{1}{r}{178.9} & \multicolumn{1}{r|}{126.7} & \multicolumn{1}{r}{263.5} & \multicolumn{1}{r|}{176.7} & \multicolumn{1}{r}{161.4} & \multicolumn{1}{r}{125.4}\\
median learning time [s] & \multicolumn{1}{r}{2.8} & \multicolumn{1}{r|}{13.4} & \multicolumn{1}{r}{0.3} & \multicolumn{1}{r|}{0.7} & \multicolumn{1}{r}{0.3} & \multicolumn{1}{r|}{1.1} & \multicolumn{1}{r}{0.3} & \multicolumn{1}{r}{0.7}\\
 -  normalized & \multicolumn{1}{r}{1.00} & \multicolumn{1}{r|}{4.85} & \multicolumn{1}{r}{1.00} & \multicolumn{1}{r|}{2.55} & \multicolumn{1}{r}{1.00} & \multicolumn{1}{r|}{3.81} & \multicolumn{1}{r}{1.00} & \multicolumn{1}{r}{2.62} \\
\bottomrule
\end{tabular}

\caption{Effect of postprocessing CPAR, CMAR, FOIL2 and PRM (base datasets) by QCBA\#5 (Q\#5 in the table). Aggregate results for 22 datasets from Table~\ref{tbl:datasets}.}

 \label{tbl:olderBench}
\end{table}

\vspace{-2mm}\subsection{Head-to-head scalability analysis}
\label{ss:scalability}
In this section, we focus on the effect of increasing the training set size on the performance of the proposed methods.
As the input dataset, we use the intrusion detection dataset from KDD Cup'99.\footnote{\url{https://www.kdd.org/kdd-cup/view/kdd-cup-1999/Data}} This dataset is suitable for benchmarking since it contains a binary class label required by the SBRL algorithm and it also contains multiple numerical attributes with a high number of distinct numerical values.  We generate subsets of the dataset containing 1,000, 10,000, 20,000, 30,000 and 40,000 rows, which is roughly the maximum number of rows for which we are able to obtain results for all implementations involved in this benchmark given the available hardware.

\begin{figure}
\centering
\begin{subfigure}{0.33\linewidth}
  \centering
  \includegraphics[height=3cm,width=\linewidth]{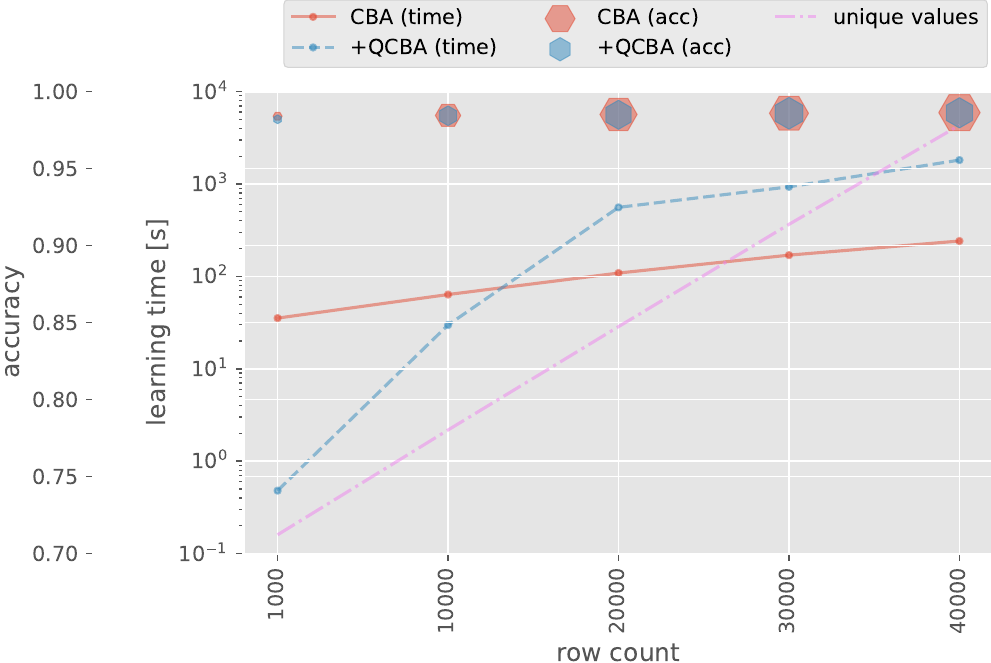} 
  \caption{CBA+QCBA}
\label{fig:sub1}
\end{subfigure}
\begin{subfigure}{0.32\linewidth}
  \centering

  \includegraphics[height=3cm,width={\dimexpr\linewidth-0.5cm\relax}]{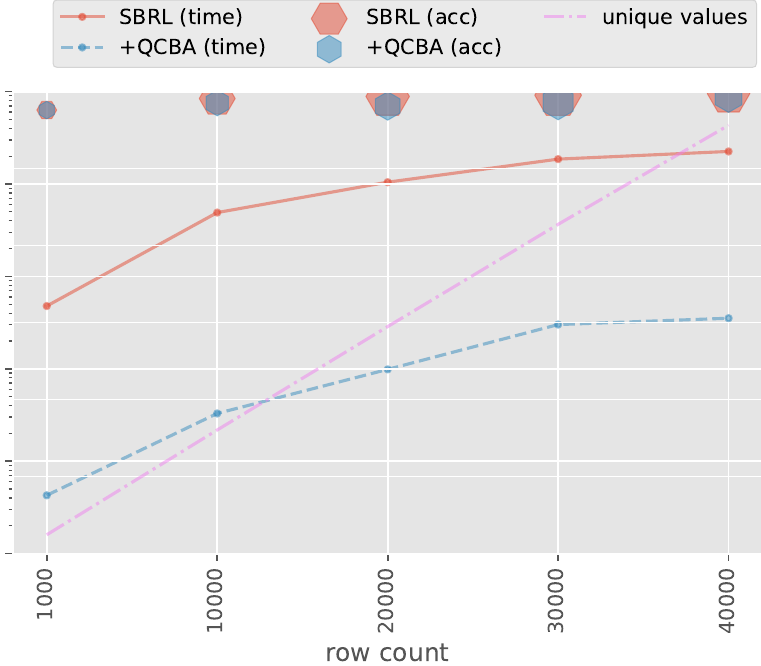} 
  \caption{SBRL+QCBA}
\label{fig:sub2}
\end{subfigure}
\begin{subfigure}{0.33\linewidth}
  \centering
  \includegraphics[height=3cm,width=\linewidth]{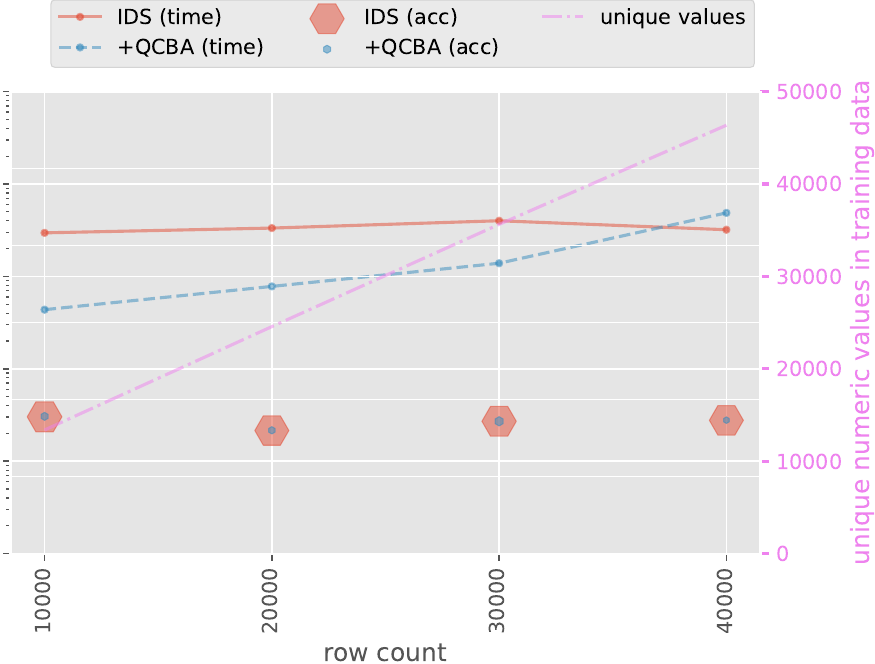} 
  \caption{IDS+QCBA}
\label{fig:sub3}
\end{subfigure}

\caption{Postprocessing on differently sized subsets of the KDD'99 anomaly dataset.  The size of the hexagonal symbol denotes the average model size.}
\label{fig:kdd}
\end{figure}

The individual subsets are preprocessed in the same way as the main evaluation on the 22 UCI datasets reported above.  Data is divided into training and testing partitions using stratified tenfold crossvalidation and the generated partitions are materialised to ensure that all evaluated configurations are compared on exactly the same data.\footnote{With one exception: result for SBRL for 40,000 rows in Figure~\ref{fig:sub1} is based on results of a single fold rather than average over all ten folds.} Numerical attributes are discretised with MDLP.

\begin{figure}
         \centering
     \includegraphics[width=\textwidth]{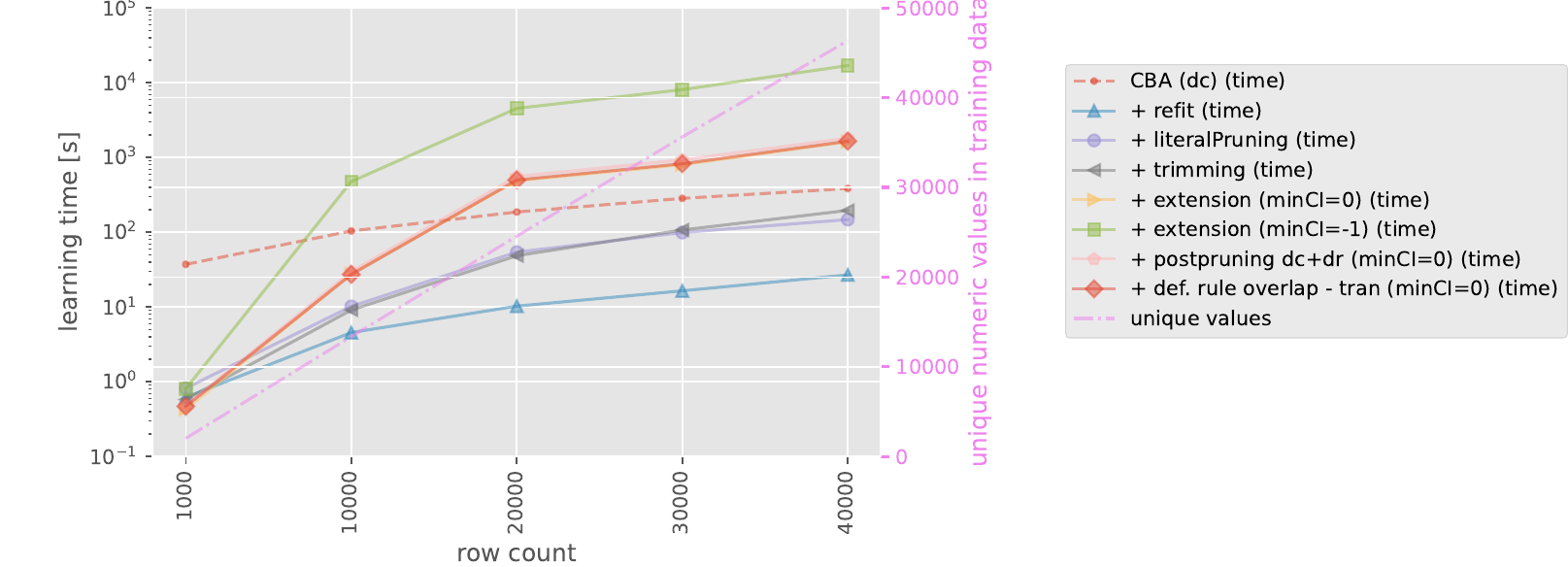} 
     \caption{Ablation study of training time for individual tuning steps in QCBA applied on a CBA model for subsets of the KDD dataset. }
     \label{fig:indivallKDDablation}
 \end{figure}

The first analysis focuses on the effect of the training set size on the time required for applying the tuning steps in QCBA. Similar to Table~\ref{tbl:qcba}, this analysis is conceived as an ``ablation study,'' where the effect of individual tuning steps on the training time is reported cumulatively. The results are included in
Figure~\ref{fig:indivallKDDablation}. The baseline is for the CBA data coverage pruning (without default rule pruning). The reported QCBA times are exclusive of CBA training time but inclusive of all preceding QCBA steps, e.g., the times for trimming also include literal pruning and refitting. Two alternative versions of the extension step are shown differing in whether a conditional accept is  enabled ($minCI=-1$) or not ($minCI=0$). The results show that conditional accept substantially increases the run time. There is, however, no effect on  accuracy (not shown in the figure). We, therefore, opted  for $minCI=0$ for the remaining experiments on the KDD Cup '99 dataset. The additional time added by post-pruning and default rule overlap pruning is small compared to the computational costs of the extension step.

A subsequent analysis summarised in Figure~\ref{fig:kdd} focuses on the effect of postprocessing models generated by three baseline algorithms by QCBA (\#5, $minCI=0$).
The settings of the baselines are similar to those in the previous experiments. For the SBRL, the maximum rule length has to be limited to 3; otherwise, the SBRL would not finish. 
 
The results indicate that postprocessing has the largest effect on the IDS, while the best results are obtained for the SBRL. Additionally,  for the SBRL, QCBA consistently reduces the model size while maintaining accuracy.  
In our evaluation, the extension step becomes computationally very intensive at approximately 40,000 rows and 40,000 distinct values. Since the used dataset contains nearly 500 thousand rows, the evaluation can be extended in future work to track performance improvements.

\vspace{-2mm}\subsection{Comparison with Related Symbolic Classifiers}
\label{ss:accuracybench}
In this section, we compare the the predictive performance against several additional reference algorithms. 

\paragraph{The selection of reference}
Algorithms C4.5 \cite{DBLP:books/mk/Quinlan93} and RIPPER \cite{Cohen:1995:FER:3091622.3091637} are well-established interpretable classifiers that are widely used as standard reference algorithms. FURIA \cite{huhn2009furia}, also covered in our related work section, is a state-of-the-art association rule classifier that outputs fuzzy rules. PART \cite{frank1998generating} is a rule learning algorithm designed to address some of the shortcomings of C4.5 and RIPPER. Despite the fact that PART is an older algorithm, it is still often used in benchmarks of recently proposed rule learning methods, such as CORELS, which is also included in the benchmark.  As the QCBA baseline, we choose the instance-based version of the default rule overlap pruning (configuration QCBA\#5 from Table~\ref{tbl:qcba}).

\vspace{-2mm}
\paragraph{Implementations}
For standard learners (C4.5, FURIA, PART, RIPPER), we use the implementations available in the Weka framework.\footnote{\url{http://www.cs.waikato.ac.nz/ml/weka/}).} 
For CORELS, we use the implementation from the authors.\footnote{Used via PyCorels Python binding \url{https://github.com/corels/pycorels}} 

\paragraph{Metaparameter Tuning}
For C4.5, FURIA, PART, and RIPPER  we perform hyperparameter optimisation using the MultiSearch package\footnote{\url{https://github.com/fracpete/multisearch-weka-package}}, which implements a variation of grid search over multi-parameter spaces. We choose accuracy as the evaluation measure.  Parameter tuning is performed separately for each fold on the training data using internal crossvalidation. The best setting found is used for the evaluation of the test data. 
The grid for individual learners is defined in Table~\ref{tbl:multisearchSetting}. 
For CORELS, it is reported that metaparameter tuning has a limited effect \cite{corelsThesis}. Similar to CBA, we  use the default values.
In this case, we use the settings from pycorels: 
{\footnotesize
\texttt{c=0.01,n\_iter = 10000,
 policy=lower\_bound,  ablation=NA,   max\_card = 2, min\_support=0.01}}. Note that minimum support is set to the same value as for CBA. For CORELS, we experimented with varying the regularisation threshold $c$ for which higher values should penalize longer rule lists, but this had only small effect, therefore the default value was used for the experiments.

 \begin{table}[htb]

 	 \centering
 {\footnotesize
 
  \centering
   \begin{minipage}{.5\linewidth}
    \centering
   \begin{tabular}{lll}
   \toprule
 parameter & value set & step\\
 \midrule
  \multicolumn{3}{c}{J48}\\
  \midrule
 confidenceFactor & [0.05,0.4] & 0.05 \\
 minNumObj & [2,14] & 4 \\
 numFolds & [2,4] & 1 \\
 reducedErrorPruning & boolean &  \\
 subtreeRaising & boolean & \\
 binarySplits & boolean & \\
 
 \midrule
 \multicolumn{3}{c}{FURIA}\\
 \midrule
 minNo & [0,1] & 0.25 \\
 folds & [2,4] & 1 \\
 
 \midrule
 \end{tabular}
 \end{minipage}%
  \begin{minipage}{.5\linewidth}
   \centering
 \begin{tabular}{lll}
   \toprule
 parameter & value set & step\\
 \midrule
 \multicolumn{3}{c}{PART}\\
 \midrule
 confidenceFactor & [0.05,0.4] & 0.05 \\
 minNumObj & [2,14] & 4 \\
 numFolds & [2,4] & 1 \\
 useMDLcorrection & boolean & \\
 reducedErrorPruning & boolean & \\
 unpruned & boolean & \\
 binarySplits & boolean & \\
 
 \midrule
 \multicolumn{3}{c}{RIPPER}\\
 \midrule
 numFolds & [2,4] & 1 \\
 checkErrorRate & boolean &  \\
 usePruning & boolean & \\
 
 \bottomrule
 
 \end{tabular}
 \end{minipage}
 }
 
 \caption{Grid definition for hyperparameter optimisation.}

 \label{tbl:multisearchSetting}
 \end{table}

\paragraph{Results}

The comparison against non-ARC classification algorithms is presented in Table~\ref{tbl:relatedlearners}.  For CORELS only the 11 datasets with binary targets are processed. For FURIA, one dataset (``letter'') is omitted due to excessive run time (the algorithm has not finished in several hours). The results show that the CBA postprocessed by the proposed tuning steps does not  have a statistically significantly better accuracy ($p<0.05$). 
However, QCBA wins on more datasets  than C4.5 (J48) and RIPPER and has the same number of wins and losses as PART. 
QCBA outperforms CORELS on all datasets except two; however, it should be noted that CORELS produces extremely small models typically composed of  only a few short rules. 
The only algorithm that, in a pairwise comparison, performs better than QCBA  is FURIA, which is a rule classifier extending on RIPPER but generating fuzzy rule models. 
A detailed discussion of the similarities and the differences between the proposed approach and FURIA and CORELS is presented in the following section.

\begin{table}[h!]
\centering

\footnotesize{
\begin{tabular}{lp{1.5cm}p{1.5cm}p{1.5cm}p{1.5cm}p{1.5cm}p{1.5cm}}
\toprule
reference classifier & CORELS & J48&PART & RIPPER  &  FURIA\\
\midrule
accuracy & 0.81 &0.84&0.85&0.79&0.85\\
won/tie/loss (CBA+QCBA vs reference) & 9/1/1   &12/2/8  & 10/3/8 & 12/5/5 & 6/4/11\\
p-value (CBA+QCBA vs reference) & 0.05  & 0.40 & 0.51 & 0.12 & 0.22 \\
\bottomrule
\end{tabular}
}
\caption{Comparison between CBA postprocessed by QCBA (\#5) and five  reference classifiers.}
\label{tbl:relatedlearners}
\end{table}

\vspace{-2mm}\subsection{Analysis of wins by dataset}
Table~\ref{tbl:wonmatrix} shows the results of individual classifiers and their combinations with QCBA per individual datasets.  The entries are sorted by the QCBA wins column, which indicates the percentage of results where QCBA has a higher accuracy in the pairwise comparisons (NA results were excluded). The last two columns show the overall best algorithm and its accuracy. Entries with the highest result obtained by QCBA are in bold. The first part of the table shows on which datasets  the postprocessing by QCBA\#5 results in an improvement over seven reference classification algorithms. The second part of the table compares five other reference classification algorithms for which postprocessing by QCBA is not available to CBA+QCBA\#5. 
Overall, QCBA was better in the majority of comparisons against the reference method on 12 datasets. QCBA combined with a baseline method obtained the best overall accuracy on five datasets.
FOIL2 is with two datasets with the highest accuracy the most successful method for combination with QCBA. The overall heterogeneity of 
the results indicates that the best algorithm or a combination of algorithms is dependent on the dataset. 
\label{ss:analysisWins}
\begin{table}[]
\centering
\begin{tabular}{lrrrrrrrrrrrrrrrl}
\toprule
\multicolumn{1}{c}{} & \multicolumn{7}{p{3cm}|}{wins of base+QCBA over base} & \multicolumn{5}{p{3cm}}{wins of CBA+QCBA over reference }&\multicolumn{1}{|p{1cm}|}{QCBA \% wins }&\multicolumn{2}{c}{best result for dataset}\\
dataset       & 1 & 2 & 3 & 4 & 5 & 6 & \multicolumn{1}{c|}{7} & 8 &9 & 10 & 11 &  12 & \multicolumn{1}{c|}{} &  accuracy &  \multicolumn{1}{c}{algorithm} \\
\midrule
labor         & 1 & 1 & 1 & 1 & 1 & 1 & NA & 1  & NA & 1 & 1  & 1 & 100 & 0.927 & \multicolumn{1}{l}{\textbf{CPAR+QCBA}}  \\
australian    & 1 & 0 & 0 & 1 & 1 & 0 & 1  & 1  & 1  & 1 & 1  & 1 & 75  & 0.868 & \multicolumn{1}{l}{CMAR}                 \\
credit-g      & 0 & 1 & 1 & 0 & 1 & 1 & 0  & 1  & 1  & 1 & 1  & 1 & 75  & 0.762 & \multicolumn{1}{l}{CBA}                  \\
letter        & 1 & 1 & 1 & 1 & 1 & 1 & NA & NA & NA & 0 & NA & 0 & 75  & 0.879 & \multicolumn{1}{l}{J48}                  \\
vehicle       & 1 & 1 & 1 & 1 & 1 & 1 & NA & 0  & NA & 0 & 0  & 1 & 70  & 0.729 & \multicolumn{1}{l}{PART}                 \\
diabetes      & 1 & 1 & 1 & 0 & 0 & 0 & 0  & 1  & 1  & 1 & 1  & 1 & 67  & 0.770 & \multicolumn{1}{l}{\textbf{CBA+QCBA}}   \\
hepatitis     & 1 & 0 & 1 & 0 & 0 & 0 & 1  & 1  & 1  & 1 & 1  & 1 & 67  & 0.832 & \multicolumn{1}{l}{CMAR}                 \\
sonar         & 1 & 0 & 1 & 1 & 0 & 1 & 0  & 0  & 1  & 1 & 1  & 1 & 67  & 0.793 & \multicolumn{1}{l}{FURIA}                \\
iris          & 1 & 1 & 0 & 1 & 1 & 0 & NA & 1  & NA & 0 & 1  & 0 & 60  & 0.960 & \multicolumn{1}{l}{\textbf{CMAR+QCBA}}  \\
lymph         & 0 & 0 & 0 & 1 & 1 & 1 & NA & 0  & NA & 1 & 1  & 1 & 60  & 0.866 & \multicolumn{1}{l}{FURIA}                \\
vowel         & 1 & 1 & 1 & 1 & 1 & 1 & NA & 0  & NA & 0 & 0  & 0 & 60  & 0.833 & \multicolumn{1}{l}{J48}                  \\
breast-w      & 0 & 0 & 1 & 1 & 1 & 1 & 1  & 0  & 1  & 1 & 0  & 0 & 58  & 0.969 & \multicolumn{1}{l}{CMAR}                 \\
autos         & 0 & 0 & 1 & 1 & 1 & 1 & NA & 0  & NA & 0 & 0  & 1 & 50  & 0.819 & \multicolumn{1}{l}{\textbf{FOIL2+QCBA}} \\
glass         & 0 & 0 & 1 & 0 & 1 & 1 & NA & 0  & NA & 1 & 0  & 1 & 50  & 0.724 & \multicolumn{1}{l}{CBA}                  \\
hypothyroid   & 1 & 1 & 1 & 1 & 0 & 1 & NA & 0  & NA & 0 & 0  & 0 & 50  & 0.996 & \multicolumn{1}{l}{J48}                  \\
heart-statlog & 0 & 0 & 0 & 1 & 0 & 0 & 1  & 0  & 1  & 1 & 0  & 1 & 42  & 0.833 & \multicolumn{1}{l}{FOIL2}                \\
ionosphere    & 0 & 0 & 0 & 0 & 1 & 0 & 1  & 0  & 1  & 1 & 1  & 0 & 42  & 0.929 & \multicolumn{1}{l}{PRM}                  \\
anneal        & 0 & 0 & 1 & 0 & 0 & 0 & NA & 0  & NA & 1 & 1  & 1 & 40  & 0.993 & \multicolumn{1}{l}{\textbf{FOIL2+QCBA}} \\
segment       & 0 & 1 & 1 & 0 & 1 & 1 & NA & 0  & NA & 0 & 0  & 0 & 40  & 0.969 & \multicolumn{1}{l}{FURIA}                \\
spambase      & 0 & 1 & 0 & 0 & 1 & 1 & 0  & 0  & 1  & 0 & 0  & 0 & 33  & 0.937 & \multicolumn{1}{l}{PART}                 \\
colic         & 0 & 0 & 1 & 0 & 0 & 1 & 1  & 0  & 0  & 0 & 0  & 0 & 25  & 0.859 & \multicolumn{1}{l}{CORELS}               \\
credit-a      & 0 & 0 & 0 & 1 & 0 & 0 & 1  & 0  & 0  & 0 & 0  & 0 & 17  & 0.867 & \multicolumn{1}{l}{CMAR}\\

\bottomrule
\end{tabular}
\caption{Results by individual datasets. Baseline and reference classifiers are coded as: 1:  CBA, 2: CMAR, 3: CPAR, 4: IDS, 5: FOIL, 6: PRM, 7: SBRL, 8: CORELS, 9: J48, 10: PART, 11: RIPPER and 12: FURIA.
}
\label{tbl:wonmatrix}
\end{table}
\section{Related Work}
\label{ss:relatedwork}
Separate-and-conquer is possibly the most common  approach to rule learning.
This strategy finds rules that explain part of training instances, separates these instances, and iteratively uses the
remaining examples to find additional rules until no instances remain \cite{furnkranz1999separate}. 
Separate-and-conquer provides a basis, for example, for the seminal RIPPER algorithm, its fuzzy rule extension with FURIA \cite{huhn2009furia}, or GuideR \cite{sikora2019guider}, a state-of-the-art algorithm allowing the user's preferences to be introduced into the mining process. The numerical attributes in separate-and-conquer approaches are typically supported through selectors ($\neq,\leq,\geq$) and range operators (intervals).
Multiple extensions to separate-and-conquer approaches have been proposed. Among them, a notable strategy is boosting (weighted covering strategy) \cite[p.175,178]{jf:Book-Nada}.  

Association rule classification is a principally different approach where many rules are first generated with fast association rule learning algorithms, and a subset of these rules is then selected to form the final classifier.
To our knowledge, all previously proposed association rule classification algorithms  support only categorical inputs. 
However, there has been some work on learning standard association rules or subgroups (as a nugget discovery rather than a classification task) from numerical data. Additionally, quantitative information is processed in fuzzy association rule classification approaches.
In the following, we discuss the differences between the new approach to supporting quantitative attributes presented in this article, which is based on postprocessing of already learned ARC models, and existing approaches for quantitative association rule learning (including those derived from formal concept analysis), fuzzy rule classification, and sequential covering (separate and conquer).

\vspace{-2mm}\subsection{Quantitative Association Rule Mining}
Several quantitative association rule learning\footnote{There is a difference between (quantitative) association rule mining, which is an exploratory data mining task aimed at discovering interesting patterns in data, and (quantitative) association rule classification, which aims to build understandable classification (predictive) models.}
 algorithms have been proposed (see \cite{adhikary2015trends} for a recent review). Two representative and widely referenced approaches include QuantMiner \cite{salleb2007quantminer} and NAR-Discovery \cite{song2013discovering}. 
The earlier proposed  QuantMiner is an evolutionary algorithm that optimises a multi-objective fitness function that combines support and confidence. The essence of QuantMiner is that multiple seed rules are subject to standard evolutionary operators. For example, mutation corresponds to an increase or a decrease in the lower/upper bound of a rule. 
NAR-Discovery takes a different, two-stage approach. Similar to QCBA, a set of ``coarse'' association rules is generated on prediscretised data, with standard association rule generation algorithms in the first stage. In the second stage, for each coarse-grained rule, several rules are generated using fine bins. The granularity of the bins is a parameter of the algorithm. One feature of NAR-Discovery is that it produces at least one order of magnitude more rules than QuantMiner.

 \begin{table}[htbp]

 {\footnotesize
 \begin{center}
 \begin{tabular}{p{3.5cm}p{2cm}p{2cm}p{2cm}}
 \toprule
 property & QCBA  & NAR-D& QuantMiner \\
 \midrule
 classification models & y & n & n\\
 deterministic & y & y & n\\
 number of rules& +++ & + & ++ \\
 precision of intervals& +++ & + & ++ \\
 externally set parameters & +++ & ++& +\\
 \bottomrule
 \end{tabular}
 \end{center}
 }
   \caption{Comparison with methods for quantitative association rule generation.}
 \label{table:comparisonQARC}

 \end{table}
 
 Table~\ref{table:comparisonQARC} compares our QCBA framework with NAR-Discovery and QuantMiner. Justifications for individual values in the table include:
 \begin{inparaenum} 
  \item classification models: neither QuantMiner or NAR-Discovery were designed for classification;
  \item deterministic: QuantMiner is an evolutionary algorithm;
  \item number of rules: too many rules generated is one of the biggest issues facing association rule generation algorithms. Neither NAR-Discovery nor QuantMiner contains procedures for limiting the number of rules, while QCBA contains several rule pruning algorithms;
  \item precision of intervals: for QuantMiner the precision  of the intervals depends on the setting of the evolutionary process and for NAR-Discovery on the discretisation setting. QCBA generates interval boundaries exactly corresponding to values in the input continuous data;
  \item externally set parameters (in addition  to confidence and support thresholds): NAR-Discovery requires two granularity settings (fine/coarse), and QuantMiner requires multiple parameters such as population size, mutation, and crossover rate for the evolutionary process, which have a considerable effect on the processing result. QCBA does not require any externally set parameters except for several optional parameters that can speed up the algorithm in exchange for a lower accuracy of the generated models.
 \end{inparaenum}
 The notion of trimming closely corresponds to the closure operator introduced in \cite{kaytoue2011revisiting}. 
Finally, it should be noted that quantitative association rule learning algorithms  address a different task than QCBA does. Unlike QCBA these are unsupervised algorithms that do not have the class information available. This is exploited by QCBA, among others, to perform rule pruning, which allows reducing the number of rules in an informed way.

\vspace{-2mm}\subsection{Numerical Pattern Mining with Formal Concept Analysis }
One approach in this area are algorithms adapting Formal Concept Analysis for the discretisation of data for pattern mining. In \cite{kaytoue2011revisiting}, an interval pattern was defined as an M-dimensional vector of intervals that could be represented as a hyperrectangle. The concept of an interval pattern is closely related to our definition of a rule as visualised in Figure~\ref{fig:all} with the difference that the problem addressed by \cite{kaytoue2011revisiting} was unsupervised. The essence of the approach in \cite{kaytoue2011revisiting} is that it finds all minimum-length interval patterns that can contain the given set of instances. These interval patterns are  called closures in \cite{kaytoue2011revisiting}. They essentially correspond to a rule after our trimming operation (Figure~\ref{fig:all}f corresponds to a closure). As a dual notion, the same article \cite{kaytoue2011revisiting} introduced the notion of a generator, which is the largest region covering the same set of instances. While we do not explicitly work with generators, they are output by QCBA as part of the extension operation (Figure~\ref{fig:all}g corresponds to a generator). The primary difference between our approach and \cite{kaytoue2011revisiting} is that the latter's goal was to create all possible closures (generators) within a given dataset, while we use the corresponding notions as a component of our approach focused on processing individual already discovered rules. According to \cite{van2016discovering}, approaches based on Formal Concept Analysis do not scale to large data. As seen from our evaluation results, by focusing on pre-mined rules we have largely (but not completely) circumvented the performance issues this approach can bring.

\vspace{-2mm}\subsection{Subgroup discovery}
Subgroup discovery is an alternative method for pattern mining. Similar to association rule mining, subgroup discovery has developed a range of methods for dealing with numerical inputs. These include discretisation of input data, as well as various heuristic approaches. A recent advance in this field is the RefineAndMine algorithm presented in \cite{belfodil2018anytime}. This algorithm  has the \emph{anytime} property, which means that its search converges to an exhaustive search if given sufficient time; however, at any time before this, a solution is available. Its main principle is somewhat similar to the NAR-Discovery quantitative association rule learning algorithm described earlier; it  first performs coarse discretisation, which is iteratively improved, and patterns are mined. However, unlike NAR-Discovery, RefineAndMine provides guarantees on the remaining distance to the exhaustive search. Mining for subgroups with numerical target attributes (also with guarantees) was addressed  in~\cite{lemmerich2016fast}. These approaches are conceptually different from ours in two respects. First, these are primarily nugget-mining algorithms intended for the discovery of a diverse set of patterns, not for generating concise classifiers. Second, these are ``all-in-one'' algorithms, while QCBA aims at postprocessing models generated by existing approaches. 

\vspace{-2mm}\subsection{Fuzzy Approaches}
Several ARC approaches  adopt fuzzy logic to deal with numerical attributes. A notable representative of this class of algorithms is FARC-HD, and its evolved version FARC-HD-OVO \cite{farchd2}. In the following, we will focus on the  FURIA algorithm \cite{huhn2009furia}. This is not an ARC algorithm but is conceptually closer to our approach and is frequently referenced as the state-of-the-art in terms of the accuracy a rule learning algorithm can achieve \cite{palacios2016extension,barsacchi2017multi}. In our benchmark, FURIA obtained the best results.

FURIA postprocesses rules generated by the RIPPER  algorithm for induction of crisp rules. RIPPER produces ordered rule lists, and the first matching rule in the list is used for classification. A default rule is added to the end of the list by RIPPER, ensuring that a query instance is always covered. As is typical for fuzzy approaches, FURIA outputs  an unordered rule set, where multiple rules may need to be combined to obtain the classification. 

A biproduct of the transition to the rule set performed by FURIA is the removal of the default rule. To ensure coverage of each instance, FURIA implements \emph{rule stretching}, which stores multiple generalisations (i.e., with one or more literals in the antecedent omitted) of each rule and uses these for classification. 
 
The most important element in  FURIA is the fuzzification of input rules.  The original intervals on quantitative attributes in rules produced by RIPPER are used as upper and lower bounds of the core $[\phi_i^{c,L}$,$\phi_i^{c,U}]$, and FURIA determines the optimal upper and lower  bounds of the fuzzy supports: $[\phi_i^{s,L}$,$\phi_i^{s,U}]$. It should be noted that in fuzzy set theory, the support bound denotes an ``upper'' rule boundary, while in the scope of association rule learning it denotes the number or fraction of correctly classified instances. When searching for $\phi_i^{s,L}$ and $\phi_i^{s,U}$, FURIA proceeds greedily; it evaluates fuzzifications for all antecedents and fuzzifies the one that produces the best purity. As fuzzification progresses, the training instances are gradually removed; therefore, the order that the rules are fuzzified in is important.

To compare FURIA with QCBA, both algorithms postprocess input rules, adjusting boundaries of numerical attributes. Within the algorithms, there are numerous differences. In QCBA, we retain the default rule (although it may be updated); therefore, even if no other rule matches, a model postprocessed with QCBA is able to provide a classification. A somewhat similar procedure to fuzzification in FURIA is the QCBA extension. Unlike in FURIA, rules are extended independently of one another, which eases parallelisation.
 In summary, FURIA produces fuzzy rules,  the resulting models are rule sets, and QCBA produces rule lists composed of crisp rules. 
 
The results for another state-of-the-art algorithm, FARC-HD, have shown that this fuzzy rule learning approach is much slower than crisp rule learning approaches. In reference to Table~\ref{tbl:comparison}, FARC-HD was on average more than 100x slower than CBA \cite{farchd2}. The remaining two other fuzzy associative classifiers (LAFAR \cite{hu2003finding} and CFAR \cite{chen2008building}) included in the benchmark in \cite{farchd2} are reported to have even more severe scalability problems as they could not be run on all datasets in the benchmark. 

For explainability (interpretability), we were unable to find a study that would evaluate the interpretability of automatically learned fuzzy rule classifiers. However, in principle, fuzzy rule algorithms generate rule sets as multiple rules are combined using membership functions to classify one instance, while models generated by QCBA are intended to be used as rule lists with only the highest-ranked rule being used to classify an instance. Rule lists and rule sets have clearly different properties in terms of interpretability, and future research will show for what purposes each representation is most suitable for.

\vspace{-2mm}\subsection{Non-Fuzzy ARC Algorithms}
In the following, we focus on the comparison of CBA (and QCBA) with  CORELS  \cite{angelino2017learning}, which is a state-of-the-art algorithm belonging to the ARC family.
We describe how CORELS works, compare it with CBA and QCBA,  discuss its performance in our experiments, and finally provide a  scalability comparison.

\vspace{-4mm}
\paragraph{CORELS - Certifiably Optimal Rule Lists}
The CORELS algorithm supports only binary class datasets, which limits the scope of comparison reported  in our evaluation only to datasets with a binary class attribute. Similar to QCBA, CORELS processes rules generated by standard association rule mining algorithms. CORELS uses a branch-and-bound method to choose a subset of the rule antecedents, transforming them into a rule list. CORELS guarantees that considering only the input rule list, the generated classifier (rule list) is optimal, where the `optimality' is defined  through a compound objective function $R$ consisting of a misclassification error on the training data (lower is better) and a regularisation term derived from a number of rules in the rule list (lower is better). When the regularisation strength is set to zero, CORELS guarantees that the produced rule list has the highest accuracy of all rule lists that can be generated by selecting antecedents from the input list of pre-mined rules. 

\vspace{-4mm}
\paragraph{Comparison of CORELS and CBA}
For rule pruning and sorting,  QCBA uses the CBA algorithm, which is based on a combination of heuristics. Rules are first sorted by their confidence, support and length and then considered in this order for addition to the final rule list. In contrast, CORELS does not use these sorting heuristics. Instead, it  performs a branch-and-bound search. Starting with an empty rule list, it iteratively finds the best rule (generated from the list of antecedents of pre-generated rules) to be added to the final rule list. CORELS exploits several properties to effectively search this very large state space. The key property  is called the \emph{hierarchical lower bound} by the CORELS authors. If the currently considered rule list (without a default rule)  $d_p$ has the same or higher misclassification error than  the objective value $R$   of the best rule list generated thus far, then the rule list $d_p$ and all extensions of it generated by appending rules at the end of $d_p$ can be omitted from the search. 
The \emph{hierarchical lower bound} property is very close to the \emph{default rule pruning} already used in CBA (see Section~\ref{ss:drpruning}).

What element of the CORELS algorithm can be attributed to producing much shorter rule lists than CBA is a matter of future research. We hypothesise that the main factor may be the replacement of the `confidence-support-rule length' sorting heuristics (see Definition~\ref{def:ranking}) used by CBA by the branch-and-bound method in CORELS. This is supported by prior research, which shows that this sorting step in CBA could be made more efficient  \cite{10.1007/978-3-540-73499-4_26}. 

\vspace{-4mm}
\paragraph{Comparison of CORELS and QCBA}
CORELS  treats the individual rule antecedents generated by rule mining as atomic (unmodifiable) components. Since CORELS is intended only for categorical data, its application on datasets with numerical attributes requires prediscretisation.
In contrast, QCBA primarily focuses on the optimisation of the individual candidate rules, editing the rules, as well as the value sets of individual literals.

\vspace{-4mm}
\paragraph{Why CORELS -- despite its optimality guarantee -- underperforms other rule learners?}
 The results that we obtain for CORELS are somewhat unexpected, but they are congruent with a detailed evaluation of CORELS performed in \cite{corelsThesis}, where CORELS was also outperformed by both PART and RIPPER, but it produced the smallest models. It should be noted that this difference cannot be accounted for a user-set bias for concise models allowed by CORELS, since setting the regularisation penalty to zero did not consistently increase the accuracy of the CORELS models.

As follows from the principles of the CORELS operation laid out in the previous, we hypothesise that the reason why rule lists generated by other rule learning algorithms  often  had higher accuracy than the ``certifiably optimal'' CORELS rule lists is that the CORELS guarantee of optimality applies only to the original candidate set of antecedents pre-mined with association rule learning from prediscretised data. This applies not only to PART and RIPPER, which do not rely on pre-mined rules but also to QCBA, which uses pre-mined rules but is able to edit these rules, which are atomic for CORELS. Thus, QCBA may work with ``better'' rules than are available to CORELS for selection to the final rule list.

\vspace{-0.5cm}
\subsection{Separate-and-conquer and rule ensembles}

We expect that state-of-the-art rule ensemble algorithms derived from RIPPER would outperform algorithms generating rule lists, even after they were postprocessed by QCBA. For example, a comparison of results published for ENDER \cite{dembczynski2010ender} with our results on several UCI datasets indicates that ENDER has higher predictive accuracy on most datasets. However,  there is an interpretability-accuracy trade-off, as rule sets are possibly less interpretable since they belong among ensemble algorithms, which are generally considered problematic from the explainability perspective \cite{setzu2021glocalx}. Furthermore, some algorithms producing rule ensembles tend to generate larger models (see rule count for CPAR in Table~\ref{tbl:comparison} is 5x higher than the rule count of CBA). 

\vspace{-4mm}
\paragraph{Scalability comparison: ARC algorithms vs. separate-and-conquer}
Association rule learning, which is a crucial computationally intensive step in ARC algorithms, is widely considered a highly scalable approach for large and sparse data (e.g., \cite{hastie01statisticallearning}). This is confirmed in a benchmark reported in \cite{DBLP:conf/clef/KliegrK15} comparing the performance of RIPPER (as a separate-and-conquer algorithm) and CBA on a larger dataset with a high number of distinct values, which shows that unlike CBA,  RIPPER is unable to process the complete dataset. In a benchmark of rule learning algorithms included in \cite{wang2017bayesian}, RIPPER was the only algorithm that is unable to finish on all datasets due to an out of memory problem on the connect-4 dataset with 67.000 instances and 42 attributes.
This indicates that while the scalability of association rule learning may face considerable challenges especially on datasets with a high number of unique values, the ARC approach as a whole shows promise and can outperform established separate-and-conquer rule learning algorithms such as RIPPER.

\section{Limitations}
\label{sec:limitations}
In this section, we first discuss the general disadvantages of building classifiers from pre-mined rules. After that, we proceed to discuss the known trade-offs in the proposed QCBA tuning steps.

\vspace{-2mm}
\paragraph{Building classifiers from pre-mined rules}
ARC algorithms rely on association rule mining for the provision of candidate rules (antecedents).  This can miss some rules that could improve the quality of the classifier. One of the reasons is that association rule mining only generates rules that exceed globally valid prespecified confidence and support thresholds. This is, for some purposes, a desirable property since rule confidence and support are computed from all instances in the dataset, and thus, the generated rules are valid independently of each other.  However, once placed into a rule list, these  independently learned rules cease to be independent. In a rule list, only instances uncovered by rules with higher precedence reach a particular rule. The generation of candidates through  association rule learning on all instances using globally set thresholds does not match the way rules are eventually used in the classifier. Since ARC algorithms such as CORELS and CBA use these independently learned rules, the resulting  rule lists can be suboptimal when all possible rules 
are considered.

\vspace{-2mm}
\paragraph{Order of processed literals }
On line~\ref{l:extit} in Procedure~\ref{alg:neighbourhood} (Extension), the literals are considered for extension in the order of appearance in the rule. A similar approach is taken in the literal  pruning algorithm, where the order of processed literals can also affect the result. 
Future work could explore an alternative  fully greedy approach, where all literals are first evaluated and then the best one is selected for processing. Such change should be done carefully to avoid adverse effects on the run time of the extension algorithm. For this, the benchmarks already show that it is the most time consuming step. For literal pruning, the expected benefits of a fully greedy approach seem to be limited, since in our experience it is rarely the case that more than one literal can be removed from a rule.

\vspace{-2mm}
\paragraph{Selection and ordering of rules in the post-pruning step}
For the post-pruning step, QCBA adopts the pruning algorithm from CBA. This is fast, but it does not provide optimal results, as it uses a heuristic to sort the rules. A possible solution would be to combine QCBA with CORELS, which provides a guarantee that the rule list is optimal with respect to the user-set trade-off between accuracy and the number of rules.  
Since CORELS operates on prediscretised data, postprocessing  CORELS models by QCBA to ``edit'' the pre-mined rules could yield additional reduction in the size of the generated models or improvements in performance of the CORELS models. 
A combination of QCBA and CORELS could take advantage of the modular architecture of QCBA, where individual tuning steps can be performed independently.

\vspace{-2mm}
\paragraph{QCBA default rule pruning step}
As follows from the benchmarks included in the previous section, neither version of default rule pruning constitutes a performance impediment. This  speed is at the expense of the exclusion of several checks that could remove additional redundant rules.  
In the original Default Rule Overlap Pruning algorithm (Proc.~\ref{alg:defRuleOverlapTrans} on line \ref{l:trc}),  the set $T_{r}^{corr}$ includes all instances in training data correctly classified by the pruning candidate $r$. 
However, instances covered by rules with higher precedence than $r$ can be excluded from $T_{r}^{corr}$.  Additional checks can also be introduced to 
ensure that the set of instances covered by each of the candidate clashes considers only instances reaching the respective candidate clash rule. Similar adjustments can be made to the range-based version of the algorithm.

\section{Conclusion}
This research aims to ameliorate one of the substantial limitations of association rule classification: the adherence of the rules that the classifier is composed of to the multidimensional grid created by discretisation of numerical attributes. 
Quantitative classification based on associations (QCBA) is, to the authors' knowledge, the first non-fuzzy association rule classification approach that recovers part of the information lost in prediscretisation. QCBA is not a standalone learning algorithm but rather a palette of postprocessing steps, from which some can be selected to enhance the results of an arbitrary rule learning algorithm. 

Prior research shows that learning algorithms that directly work with numerical attributes do not often have better predictive performance than when the discretisation is performed as part of preprocessing \cite{ghodke2007investigation,bryson2001attribute}. 
Our results obtained for classification rules are partly in line with the aforementioned conclusions that were obtained for decision trees and probabilistic approaches. 

As part of the evaluation, we  postprocessed models generated by CBA, CMAR, CPAR, IDS, FOIL2, PRM and SBRL  and  compared the result of the postprocessing  with the baseline version. There was an improvement in predictive accuracy or the won-tie-loss record for all baseline methods except CMAR. While the effect on predictive accuracy was typically relatively small, bigger and consistent improvements could be observed for model size. For example, baseline CMAR generated by far the largest model which the QCBA algorithm reduced by nearly 80\% without negatively affecting the predictive performance.

An interesting area of future work would be combining QCBA with a broader range of base rule learning algorithms and problems, such as multilabel classification \cite{rapp2021boomer,hullermeier2020rule}. The incorporation of a wider range of rule quality measures could also have a positive impact on predictive performance \cite{wrobel2016rule}.
Real-world datasets typically contain both numerical and categorical attributes, while the presented work addresses only quantitative attributes. There is a complementary line of research on merging the categorical attributes by creating multi-value rule sets \cite{wang2018multi}, with very promising results in terms of gains in accuracy and comprehensibility. Since QCBA has a modular architecture, some of its tuning steps can become building blocks in a combined approach that would generate small, yet accurate models on data containing mixed attribute types.

\section*{Acknowledgements}
We would like to thank the anonymous reviewers for the very detailed, thorough and insightful feedback that led to many  improvements in the manuscript.
We would also like to thank Ji\v{r}{\'i} Filip for feedback on the pseudocode listings in the early version of the  manuscript and for implementing an alternative QCBA implementation, which is available in the pyARC Python package for association rule classification. 
An earlier version of part of the text used in this article was included in the first author's PhD thesis at the Queen Mary University of London \cite{kliegr2017effect}. 

\section*{Statements and Declarations}
\paragraph{Funding}
TK was supported by the Faculty of Informatics and Statistics, VSE by institutional support for long-term research. The  scalability experiments on the KDD'99 intrusion detection dataset were supported by grant IGA 37/2021 of Prague University of Economics and Business.
\vspace{-0.5cm}
\paragraph{Conflicts of interest/Competing interests}
The authors declare no conflicts of interest and no competing interests.
\vspace{-0.5cm}
\paragraph{Ethics approval}
Not applicable
\vspace{-0.5cm}
\paragraph{Consent to participate}
Not applicable
\vspace{-0.5cm}
\paragraph{Consent for publication}
Not applicable
\vspace{-0.5cm}
\paragraph{Availability of data and material}
The datasets as well as the source code for conducting the experiments is available  at \url{https://github.com/kliegr/arcBench}.
\vspace{-0.5cm}
\paragraph{Code availability}
The implementation of the proposed algorithms in Java with R interface is available under an open licence at \url{http://github.com/kliegr/qcba} (also in the CRAN  repository as package \texttt{qCBA}).
\vspace{-0.5cm}
\paragraph{Authors' contributions}
TK design, implementation and writing. EI supervision of the research.

\bibliographystyle{spmpsci}
\bibliography{zdroje2}

\begin{thebibliography}{10}
\providecommand{\url}[1]{{#1}}
\providecommand{\urlprefix}{URL }
\expandafter\ifx\csname urlstyle\endcsname\relax
  \providecommand{\doi}[1]{DOI~\discretionary{}{}{}#1}\else
  \providecommand{\doi}{DOI~\discretionary{}{}{}\begingroup
  \urlstyle{rm}\Url}\fi

\bibitem{adhikary2015trends}
Adhikary, D., Roy, S.: Trends in quantitative association rule mining
  techniques.
\newblock In: Recent Trends in Information Systems (ReTIS), 2015 IEEE 2nd
  International Conference on, pp. 126--131. IEEE (2015)

\bibitem{agrawal1998automatic}
Agrawal, R., Gehrke, J., Gunopulos, D., Raghavan, P.: Automatic subspace
  clustering of high dimensional data for data mining applications, vol.~27.
\newblock ACM (1998)

\bibitem{DBLP:conf/sigmod/AgrawalIS93}
Agrawal, R., Imielinski, T., Swami, A.N.: Mining association rules between sets
  of items in large databases.
\newblock In: SIGMOD, pp. 207--216. ACM Press (1993)

\bibitem{alcala2011fuzzy}
Alcala-Fdez, J., Alcala, R., Herrera, F.: A fuzzy association rule-based
  classification model for high-dimensional problems with genetic rule
  selection and lateral tuning.
\newblock IEEE Transactions on Fuzzy Systems \textbf{19}(5), 857--872 (2011)

\bibitem{angelino2017learning}
Angelino, E., Larus-Stone, N., Alabi, D., Seltzer, M., Rudin, C.: Learning
  certifiably optimal rule lists for categorical data.
\newblock The Journal of Machine Learning Research \textbf{18}(1), 8753--8830
  (2017)

\bibitem{barsacchi2017multi}
Barsacchi, M., Bechini, A., Marcelloni, F.: Multi-class boosting with fuzzy
  decision trees.
\newblock In: Fuzzy Systems (FUZZ-IEEE), 2017 IEEE International Conference on,
  pp. 1--6. IEEE (2017)

\bibitem{belfodil2018anytime}
Belfodil, A., Belfodil, A., Kaytoue, M.: Anytime subgroup discovery in
  numerical domains with guarantees.
\newblock In: Joint European Conference on Machine Learning and Knowledge
  Discovery in Databases, pp. 500--516. Springer (2018)

\bibitem{benavoli2016should}
Benavoli, A., Corani, G., Mangili, F.: Should we really use post-hoc tests
  based on mean-ranks?
\newblock The Journal of Machine Learning Research \textbf{17}(1), 152--161
  (2016)

\bibitem{bryson2001attribute}
Bryson, N., Giles, K.: Attribute discretization for classification.
\newblock In: Proceedings of Americas conference on information systems (AMCIS
  2021) (2001)

\bibitem{chen2008building}
Chen, Z., Chen, G.: Building an associative classifier based on fuzzy
  association rules.
\newblock International Journal of Computational Intelligence Systems
  \textbf{1}(3), 262--273 (2008)

\bibitem{Cohen:1995:FER:3091622.3091637}
Cohen, W.W.: Fast effective rule induction.
\newblock In: Proceedings of the Twelfth International Conference on
  International Conference on Machine Learning, ICML'95, pp. 115--123. Morgan
  Kaufmann Publishers Inc., San Francisco, CA, USA (1995)

\bibitem{dembczynski2010ender}
Dembczy{\'n}ski, K., Kot{\l}owski, W., S{\l}owi{\'n}ski, R.: {ENDER}: a
  statistical framework for boosting decision rules.
\newblock Data Mining and Knowledge Discovery \textbf{21}(1), 52--90 (2010)

\bibitem{djenouri2018mining}
Djenouri, Y., Belhadi, A., Fournier-Viger, P., Fujita, H.: Mining diversified
  association rules in big datasets: a cluster/{GPU}/genetic approach.
\newblock Information Sciences \textbf{459}, 117--134 (2018)

\bibitem{farchd2}
Elkano, M., Galar, M., Sanz, J.A., Fernández, A., Barrenechea, E., Herrera,
  F., Bustince, H.: Enhancing multiclass classification in {FARC-HD} fuzzy
  classifier: On the synergy between $n$-dimensional overlap functions and
  decomposition strategies.
\newblock IEEE Transactions on Fuzzy Systems \textbf{23}(5), 1562--1580 (2015).
\newblock \doi{10.1109/TFUZZ.2014.2370677}

\bibitem{conf/ijcai/FayyadI93}
Fayyad, U.M., Irani, K.B.: Multi-interval discretization of continuous-valued
  attributes for classification learning.
\newblock In: 13th International Joint Conference on Uncertainly in Artificial
  Intelligence (IJCAI93), pp. 1022--1029 (1993)

\bibitem{feige2011maximizing}
Feige, U., Mirrokni, V.S., Vondr{\'a}k, J.: Maximizing non-monotone submodular
  functions.
\newblock SIAM Journal on Computing \textbf{40}(4), 1133--1153 (2011)

\bibitem{feng2016soft}
Feng, F., Cho, J., Pedrycz, W., Fujita, H., Herawan, T.: Soft set based
  association rule mining.
\newblock Knowledge-Based Systems \textbf{111}, 268--282 (2016)

\bibitem{fernandes2022machine}
Fernandes, M., Corchado, J.M., Marreiros, G.: Machine learning techniques
  applied to mechanical fault diagnosis and fault prognosis in the context of
  real industrial manufacturing use-cases: a systematic literature review.
\newblock Applied Intelligence pp. 1--35 (2022)

\bibitem{idsRULEML}
Filip, J., Kliegr, T.: Py{IDS}–{P}ython implementation of interpretable
  decision sets algorithm by {Lakkaraju} et al, 2016.
\newblock In: RuleML Challenge, {RuleML+RR} 2019. CEUR-WS (2019)

\bibitem{frank1998generating}
Frank, E., Witten, I.H.: Generating accurate rule sets without global
  optimization.
\newblock In: Proceedings of the Fifteenth International Conference on Machine
  Learning, ICML '98, p. 144–151. Morgan Kaufmann Publishers Inc., San
  Francisco, CA, USA (1998)

\bibitem{friedman2009elements}
Friedman, J., Hastie, T., Tibshirani, R.: The elements of statistical learning:
  Data mining, inference, and prediction.
\newblock Springer Series in Statistics  (2009)

\bibitem{furnkranz1999separate}
F{\"u}rnkranz, J.: Separate-and-conquer rule learning.
\newblock Artificial Intelligence Review \textbf{13}(1), 3--54 (1999)

\bibitem{jf:Book-Nada}
F{\"{u}}rnkranz, J., Gamberger, D., Lavra{\v c}, N.: Foundations of Rule
  Learning.
\newblock Springer-Verlag (2012)

\bibitem{furnkranz2015brief}
F{\"u}rnkranz, J., Kliegr, T.: A brief overview of rule learning.
\newblock In: International Symposium on Rules and Rule Markup Languages for
  the Semantic Web, pp. 54--69. Springer (2015)

\bibitem{furnkranz2020cognitive}
F{\"u}rnkranz, J., Kliegr, T., Paulheim, H.: On cognitive preferences and the
  plausibility of rule-based models.
\newblock Machine Learning \textbf{109}(4), 853--898 (2020)

\bibitem{ghodke2007investigation}
Ghodke, S., Baldwin, T.: An investigation into the interaction between feature
  selection and discretization: Learning how and when to read numbers.
\newblock In: Australasian Joint Conference on Artificial Intelligence, pp.
  48--57. Springer (2007)

\bibitem{giacometti2018dense}
Giacometti, A., Soulet, A.: Dense neighborhood pattern sampling in numerical
  data.
\newblock In: Proceedings of the 2018 SIAM International Conference on Data
  Mining, pp. 756--764. SIAM (2018)

\bibitem{gonzalez2001selection}
Gonz{\'a}lez, A., P{\'e}rez, R.: Selection of relevant features in a fuzzy
  genetic learning algorithm.
\newblock IEEE Transactions on Systems, Man, and Cybernetics, Part B
  (Cybernetics) \textbf{31}(3), 417--425 (2001)

\bibitem{arules}
Hahsler, M., Grun, B., Hornik, K.: Introduction to arules -- Mining Association
  Rules and Frequent Item Sets.
\newblock
  \urlprefix\url{https://cran.r-project.org/web/packages/arules/vignettes/arules.pdf}.
\newblock Accessed [1-Jan-2023]

\bibitem{arulesCBA}
Hahsler, M., Johnson, I., Giallanza, T.: arulesCBA: Classification Based on
  Association Rules (2022).
\newblock \urlprefix\url{https://CRAN.R-project.org/package=arulesCBA}.
\newblock {R} package version 1.2.4, Accessed 1-Aug-2022

\bibitem{guhaOverview}
H{\'a}jek, P., Hole\v{n}a, M., Rauch, J.: The {GUHA} method and its meaning for
  data mining.
\newblock Journal of Computer and System Sciences \textbf{76}, 34--48 (2010)

\bibitem{Han:2004:MFP:954514.954525}
Han, J., Pei, J., Yin, Y., Mao, R.: Mining frequent patterns without candidate
  generation: A frequent-pattern tree approach.
\newblock Data Mining and Knowledge Discovery \textbf{8}(1), 53--87 (2004)

\bibitem{hastie01statisticallearning}
Hastie, T., Tibshirani, R., Friedman, J.: The Elements of Statistical Learning.
\newblock Springer Series in Statistics. Springer New York Inc., New York, NY,
  USA (2001)

\bibitem{hu2003finding}
Hu, Y.C., Chen, R.S., Tzeng, G.H.: Finding fuzzy classification rules using
  data mining techniques.
\newblock Pattern Recognition Letters \textbf{24}(1-3), 509--519 (2003)

\bibitem{huhn2009furia}
H{\"u}hn, J., H{\"u}llermeier, E.: {FURIA}: an algorithm for unordered fuzzy
  rule induction.
\newblock Data Mining and Knowledge Discovery \textbf{19}(3), 293--319 (2009)

\bibitem{hullermeier2020rule}
H{\"u}llermeier, E., F{\"u}rnkranz, J., Mencia, E.L., Nguyen, V.L., Rapp, M.:
  Rule-based multi-label classification: Challenges and opportunities.
\newblock In: International Joint Conference on Rules and Reasoning, pp. 3--19.
  Springer (2020)

\bibitem{ishibuchi2005hybridization}
Ishibuchi, H., Yamamoto, T., Nakashima, T.: Hybridization of fuzzy {GBML}
  approaches for pattern classification problems.
\newblock IEEE Transactions on Systems, Man, and Cybernetics, Part B
  (Cybernetics) \textbf{35}(2), 359--365 (2005)

\bibitem{kaytoue2011revisiting}
Kaytoue, M., Kuznetsov, S.O., Napoli, A.: Revisiting numerical pattern mining
  with formal concept analysis.
\newblock In: Twenty-Second International Joint Conference on Artificial
  Intelligence - IJCAI 2011 (2011)

\bibitem{kliegr2017effect}
Kliegr, T.: Effect of cognitive biases on human understanding of rule-based
  machine learning models.
\newblock Ph.D. thesis, {Queen Mary University of London} (2017)

\bibitem{DBLP:conf/ruleml/KliegrKSV14}
Kliegr, T., Kucha{\v{r}}, J., Sottara, D., Voj{\'i}{\v{r}}, S.: Learning
  business rules with association rule classifiers.
\newblock In: A.~Bikakis, P.~Fodor, D.~Roman (eds.) Rules on the Web. From
  Theory to Applications: 8th International Symposium, RuleML 2014, Co-located
  with the 21st European Conference on Artificial Intelligence, ECAI 2014,
  Prague, Czech Republic, August 18-20, 2014. Proceedings, pp. 236--250.
  Springer International Publishing, Cham (2014)

\bibitem{DBLP:conf/clef/KliegrK15}
Kliegr, T., Kucha\v{r}, J.: Benchmark of rule-based classifiers in the news
  recommendation task.
\newblock In: CLEF Proceedings, \emph{LNCS}, vol. 9283, pp. 130--141. Springer
  (2015)

\bibitem{lakkarajuinterpretable}
Lakkaraju, H., Bach, S.H., Leskovec, J.: Interpretable decision sets: A joint
  framework for description and prediction.
\newblock In: Proceedings of the 22Nd ACM SIGKDD International Conference on
  Knowledge Discovery and Data Mining, KDD '16, pp. 1675--1684. ACM, New York,
  NY, USA (2016)

\bibitem{lemmerich2016fast}
Lemmerich, F., Atzmueller, M., Puppe, F.: Fast exhaustive subgroup discovery
  with numerical target concepts.
\newblock Data Mining and Knowledge Discovery \textbf{30}(3), 711--762 (2016)

\bibitem{letham2015interpretable}
Letham, B., Rudin, C., McCormick, T.H., Madigan, D.: Interpretable classifiers
  using rules and bayesian analysis: Building a better stroke prediction model.
\newblock The Annals of Applied Statistics \textbf{9}(3), 1350--1371 (2015)

\bibitem{li2001cmar}
Li, W., Han, J., Pei, J.: {CMAR}: Accurate and efficient classification based
  on multiple class-association rules.
\newblock In: Data Mining, 2001. ICDM 2001, Proceedings IEEE International
  Conference on, pp. 369--376. IEEE (2001)

\bibitem{Liu98integratingclassification}
Liu, B., Hsu, W., Ma, Y.: Integrating classification and association rule
  mining.
\newblock In: Proceedings of the Fourth International Conference on Knowledge
  Discovery and Data Mining, KDD'98, pp. 80--86. AAAI Press (1998)

\bibitem{liu2001classification}
Liu, B., Ma, Y., Wong, C.K.: Classification using association rules: weaknesses
  and enhancements.
\newblock Data mining for scientific applications \textbf{591} (2001)

\bibitem{mansoori2008sgerd}
Mansoori, E.G., Zolghadri, M.J., Katebi, S.D.: {SGERD}: A steady-state genetic
  algorithm for extracting fuzzy classification rules from data.
\newblock IEEE Transactions on Fuzzy Systems \textbf{16}(4), 1061--1071 (2008)

\bibitem{nawaz2021using}
Nawaz, M.S., Fournier-Viger, P., Shojaee, A., Fujita, H.: Using artificial
  intelligence techniques for {COVID-19} genome analysis.
\newblock Applied Intelligence \textbf{51}(5), 3086--3103 (2021)

\bibitem{palacios2016extension}
Palacios, A., S{\'a}nchez, L., Couso, I., Destercke, S.: An extension of the
  {FURIA} classification algorithm to low quality data through fuzzy rankings
  and its application to the early diagnosis of dyslexia.
\newblock Neurocomputing \textbf{176}, 60--71 (2016)

\bibitem{DBLP:books/mk/Quinlan93}
Quinlan, J.R.: {C4.5}: Programs for Machine Learning.
\newblock Morgan Kaufmann (1993)

\bibitem{quinlan1996improved}
Quinlan, J.R.: Improved use of continuous attributes in {C4. 5}.
\newblock Journal of artificial intelligence research \textbf{4}, 77--90 (1996)

\bibitem{quinlan1993foil}
Quinlan, J.R., Cameron-Jones, R.M.: {FOIL}: A midterm report.
\newblock In: European conference on machine learning, pp. 1--20. Springer
  (1993)

\bibitem{rapp2021boomer}
Rapp, M.: Boomer—an algorithm for learning gradient boosted multi-label
  classification rules.
\newblock Software Impacts \textbf{10}, 100137 (2021)

\bibitem{rudin2019stop}
Rudin, C.: Stop explaining black box machine learning models for high stakes
  decisions and use interpretable models instead.
\newblock Nature Machine Intelligence \textbf{1}(5), 206--215 (2019)

\bibitem{salleb2007quantminer}
Salleb-Aouissi, A., Vrain, C., Nortet, C.: Quantminer: A genetic algorithm for
  mining quantitative association rules.
\newblock In: IJCAI, vol.~7 (2007)

\bibitem{schmidhuber2015deep}
Schmidhuber, J.: Deep learning in neural networks: An overview.
\newblock Neural networks \textbf{61}, 85--117 (2015)

\bibitem{setzu2021glocalx}
Setzu, M., Guidotti, R., Monreale, A., Turini, F., Pedreschi, D., Giannotti,
  F.: {GlocalX}-from local to global explanations of black box {AI} models.
\newblock Artificial Intelligence \textbf{294}, 103457 (2021)

\bibitem{sikora2019guider}
Sikora, M., Wr{\'o}bel, {\L}., Gudy{\'s}, A.: {GuideR}: A guided
  separate-and-conquer rule learning in classification, regression, and
  survival settings.
\newblock Knowledge-Based Systems \textbf{173}, 1--14 (2019)

\bibitem{song2013discovering}
Song, C., Ge, T.: Discovering and managing quantitative association rules.
\newblock In: Proceedings of the 22nd ACM international conference on
  information \& knowledge management, pp. 2429--2434. ACM (2013)

\bibitem{corelsThesis}
Speh, C.: Evaluation of different rule learning algorithms.
\newblock TU Darmstadt (2019).
\newblock Bachelor Thesis

\bibitem{thabtah2006pruning}
Thabtah, F.: Pruning techniques in associative classification: Survey and
  comparison.
\newblock Journal of Digital Information Management \textbf{4}(3) (2006)

\bibitem{van2016discovering}
Van~Brussel, T., M{\"u}ller, E., Goethals, B.: Discovering overlapping
  quantitative associations by density-based mining of relevant attributes.
\newblock In: FoIKS, pp. 131--148. Springer (2016)

\bibitem{arcReview}
Vanhoof, K., Depaire, B.: Structure of association rule classifiers: a review.
\newblock In: 2010 International Conference on Intelligent Systems and
  Knowledge Engineering (ISKE), pp. 9--12 (2010)

\bibitem{wang2018multi}
Wang, T.: Multi-value rule sets for interpretable classification with
  feature-efficient representations.
\newblock In: Advances in Neural Information Processing Systems, pp.
  10835--10845 (2018)

\bibitem{wang2017bayesian}
Wang, T., Rudin, C., Doshi-Velez, F., Liu, Y., Klampfl, E., MacNeille, P.: A
  bayesian framework for learning rule sets for interpretable classification.
\newblock The Journal of Machine Learning Research \textbf{18}(1), 2357--2393
  (2017)

\bibitem{10.1007/978-3-540-73499-4_26}
Wang, Y.J., Xin, Q., Coenen, F.: A novel rule ordering approach in
  classification association rule mining.
\newblock In: P.~Perner (ed.) Machine Learning and Data Mining in Pattern
  Recognition, pp. 339--348. Springer Berlin Heidelberg, Berlin, Heidelberg
  (2007)

\bibitem{wrobel2016rule}
Wr{\'o}bel, {\L}., Sikora, M., Michalak, M.: Rule quality measures settings in
  classification, regression and survival rule induction—an empirical
  approach.
\newblock Fundamenta Informaticae \textbf{149}(4), 419--449 (2016)

\bibitem{yang2017scalable}
Yang, H., Rudin, C., Seltzer, M.: Scalable bayesian rule lists.
\newblock In: Proceedings of the 34th International Conference on Machine
  Learning-Volume 70, pp. 3921--3930. JMLR (2017)

\bibitem{yin2003cpar}
Yin, X., Han, J.: {CPAR}: Classification based on predictive association rules.
\newblock In: Proceedings of the 2003 SIAM International Conference on Data
  Mining, pp. 331--335. SIAM (2003)

\bibitem{zaki2000scalable}
Zaki, M.J.: Scalable algorithms for association mining.
\newblock IEEE Transactions on Knowledge and Data Engineering \textbf{12}(3),
  372--390 (2000)

\end{thebibliography}

\appendix
\clearpage
\section{Appendix}
\label{sec:properties}

The notion of density and volume within the scope of frequent itemset mining already appeared in \cite{agrawal1998automatic} as follows: the data space is divided by discretisation (equal-length intervals) into indivisible units. Each unit is assumed to have the same volume, and the density corresponds to the number of points inside the unit.
Since the purpose of our work is to dismantle the intervals created by discretisation, we need to define the rule density and the volume more precisely, taking into account the real length of intervals and addressing nominal attributes. Another more recent work focusing on defining the density of patterns is  \cite{giacometti2018dense}, who defined  density somewhat differently from us.\footnote{Article \cite{giacometti2018dense} was published in parallel to our ongoing work.}

\begin{definition}
\label{def:rulevolume}
\textbf{(rule volume)} The volume of a rule $r$ is computed as follows:
\begin{equation}
volume(r) = \prod_{A \in \mathcal{A}} length(r,A),
\end{equation}
where $\mathcal{A}$  is the set of predictor attributes in dataset $T$. The length computation for attribute $A$ depends on whether the rule contains a literal defined on attribute $A$ and if so, on the type of the value in the literal as follows:

\begin{equation}
length(r,A) = 
\begin{cases}
\frac{max(vals(V,T))-min(vals(V,T))}{max(vals(A,T))-min(vals(A,T))} & \exists A(V)\in ant(r):\mbox{ } V \mbox{ is an interval } \\
\frac{1}{|vals(A,T)|} & \exists A(V)\in ant(r): \mbox{ } V \mbox{ is a nominal value}\\
1  & \nexists (A,V)\in ant(r) \\
\end{cases} 
\end{equation}
\end{definition}

\begin{definition}
\label{def:ruledense}
\textbf{(rule density)} The density of rule $r$ is computed as follows:

\begin{equation}
\rho(r)= \frac{supp_{abs}(r)}{volume(r)}.
\end{equation}
\end{definition}

The intuition behind the density equation is that $supp_{abs}$ corresponds to the number of ``particles of interest'': instances of the class in the rule consequent that are covered by the rule. 

\begin{algorithm}
\scriptsize
\caption{Default Rule Overlap Pruning (Range-based) \emph{drop-ra()}}
\label{alg:defRuleOverlapRange}
\begin{algorithmic}[1]
\REQUIRE \emph{rules} to be pruned, $T$ training instances  described by attributes $A_1, \ldots,  A_n, C$.
\ENSURE pruned $rules$ (some elements of $rules$ removed)
\STATEx 
\STATE  $r_d \leftarrow$ default  rule in $rules$
\FORALL{ $ r \in rules \setminus r_d : cons(r) = cons(r_d)$ }\COMMENT{Iterate from the highest precedence to lowest}

    \STATE $R_{clash} \leftarrow \{r' \in rules:  r' \mbox{ has lower precedence than } r  \mbox{  and }  cons(r) \neq cons(r_d)\}$ \Comment{Candidate clashes}
    
    \IF{($\forall r'\in R_{clash}) (\{attr(l) : l\in ant(r')\} \cap \{attr(l) : l\in ant(r)\}\neq \emptyset)$} 
 \Comment{A rule $r'$ without  attribute shared with $r$ would also cover part of the same space  covered by $r$ and therefore $r$ can be pruned only if all $r'$ have at least one shared attribute with $r$. }
    \IF{$(\forall r'' \in R_{clash})(\exists A(V'')\in ant(r''), \exists A(V) \in ant(r))(V''\cap V=\emptyset)$}
  \Comment{All candidate clash rules $r''$ must share at least one  attribute with pruning candidate $r$ with disjoint values of the respective literals. }
     \STATE $rules$ $\leftarrow$ $rules \setminus r$ \Comment{None of the candidate clashes cover regions covered by the pruning candidate $r$, $r$ can be removed}
    \ENDIF
    \ENDIF
    \ENDFOR   
  \STATE  \textbf{return}  \emph{rules}

\end{algorithmic}
\end{algorithm}

\end{document}